\PassOptionsToPackage{table,dvipsnames}{xcolor}
\documentclass[a4paper,fleqn]{cas-sc}

\usepackage[numbers,sort&compress]{natbib}
\setcitestyle{numbers,square,comma,sort&compress}
\usepackage{graphicx}
\usepackage{dsfont}
\usepackage{subfig}
\usepackage{booktabs}
\usepackage{tabularx}
\usepackage{multirow}
\usepackage{makecell}
\usepackage{array}
\usepackage{threeparttable}
\usepackage{stfloats}
\usepackage{amssymb}
\usepackage{amsmath}
\usepackage{siunitx}
\usepackage{lineno}
\usepackage{longtable}
\usepackage{ragged2e}

\usepackage{xcolor}
\usepackage{pifont}
\usepackage{algorithm}
\usepackage{algpseudocode}

\usepackage{adjustbox}

\definecolor{ElsevierBlue}{HTML}{00AEEF}

\definecolor{boxaccent}{RGB}{84,100,123}
\definecolor{maskaccent}{RGB}{96,119,105}
\definecolor{contaccent}{RGB}{128,98,76}

\definecolor{bestbg}{RGB}{217,231,249}      
\definecolor{bestfg}{RGB}{24,56,92}
\definecolor{secondbg}{RGB}{236,240,244}
\definecolor{secondfg}{RGB}{88,96,108}
\definecolor{oursbg}{RGB}{236,244,234}
\definecolor{summarygray}{RGB}{245,245,245}
\definecolor{tableblue}{RGB}{232,240,255}
\definecolor{gaincolor}{RGB}{0,110,0}

\definecolor{shapebestbg}{RGB}{246,223,190}
\definecolor{shapebestfg}{RGB}{115,68,20}

\definecolor{geombestbg}{RGB}{235,231,244}
\definecolor{geombestfg}{RGB}{86,74,120}

\definecolor{SoftBlue}{RGB}{50, 90, 140}
\definecolor{SoftRed}{RGB}{180, 90, 80}
\definecolor{SoftGray}{RGB}{113, 128, 150}

\algrenewcommand\algorithmiccomment[1]{\hfill\textcolor{SoftGray}{// #1}}

\newcommand{\abyes}{\raisebox{0.10ex}{\scalebox{0.92}{\textcolor{ForestGreen!70!black}{\ding{51}}}}}
\newcommand{\abno}{\raisebox{0.10ex}{\scalebox{0.92}{\textcolor{BrickRed!80!black}{\ding{55}}}}}

\newcommand{\abdash}{\textcolor{black!55}{--}}

\AtBeginDocument{%
  \hypersetup{
    colorlinks=true,
    linkcolor=ElsevierBlue,
    citecolor=ElsevierBlue,
    urlcolor=ElsevierBlue,
  }%
}

\begin{document}

\shorttitle{Degradation-Aware Mixture-of-Experts Enhancement}
\shortauthors{H. Wang  et~al.}

\title [mode = title]{Bridge Damage Detection from Low-Light UAV Imagery via Degradation-Aware Mixture-of-Experts Enhancement}                      

\author[1]{Hu Wang}
\author[2]{Hongxu Pu}
\author[3, 4]{Zhiqi Hu}
\author[5]{Fangzhou Lin}
\author[2]{Wang Wang}
\ead{wangw00821@gmail.com}
\cormark[1]

\affiliation[1]{organization={School of Computer Science and Engineering, University of Electronic Science and Technology of China},
                addressline={No. 2006, Xiyuan Avenue, West Hi-Tech Zone}, 
                city={Chengdu},
                postcode={611731}, 
                country={China}}

\affiliation[2]{organization={Sustainability X-Lab, The University of Hong Kong},
                city={Hong Kong},
                country={China}}

\affiliation[3]{organization={School of Architecture, Building, and Civil Engineering, Loughborough University},
                city={Loughborough},
                country={UK, LE11 3TU}}

\affiliation[4]{organization={Department of Engineering Science, University of Oxford},
                city={Oxford},
                country={UK}}

\affiliation[5]{organization={Department of Civil and Environmental Engineering, The Hong Kong University of Science                and Technology}, 
               addressline={Clear Water Bay, Kowloon}, 
               city={Hong Kong}, 
               country={China}}


\begin{abstract}
Poor illumination obscures small, low-contrast defects in UAV bridge imagery, reducing the reliability and operational flexibility of automated inspection.
This paper investigates whether degradation-aware image restoration can improve bridge damage detection under low-light conditions and transfer from synthetic degradations to real inspection scenes.
We propose DaL-MoE, a detector-agnostic restoration front end trained with an ISP-aware low-light synthesis pipeline and equipped with degradation-aware guidance estimation and complementary experts for noise suppression, color adjustment, and structural-detail recovery.
On paired synthetic data, DaL-MoE achieves 23.12~dB PSNR and 0.8482 SSIM, increasing YOLOv11m box $\mathrm{mAP}_{50}$ from 0.3097 to 0.4923 and mask $\mathrm{mAP}_{50}$ from 0.2281 to 0.3529.
On real low-light UAV imagery without paired normal-light references, sim-to-real evaluation shows improved defect visibility and more complete detections than direct inference on raw low-light inputs.
Future work will develop low-light-aware bridge damage detectors with stronger cross-scene generalization across bridge sites, imaging conditions, and illumination levels.
\end{abstract}


\maketitle

\begin{figure}[pos=h]
    \centering
    \includegraphics[width=0.73\linewidth]{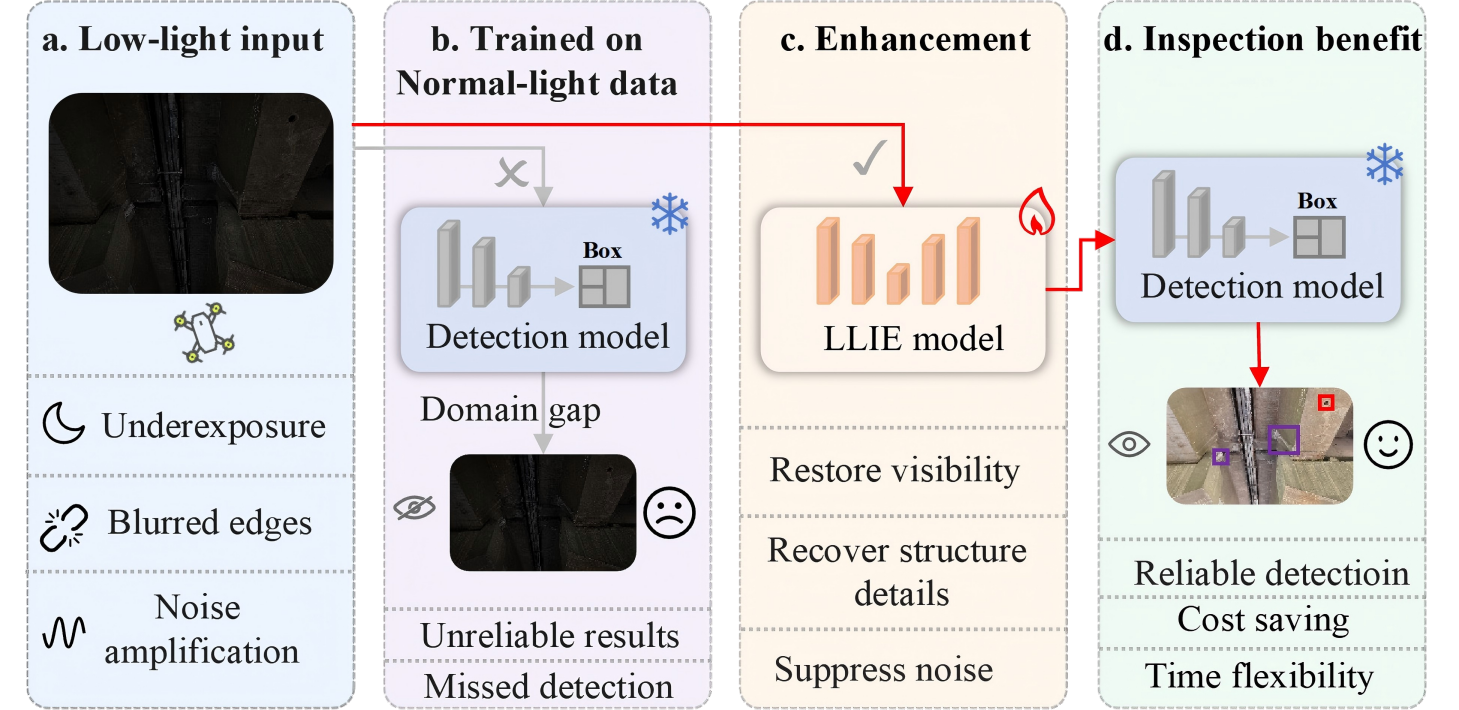}
    \caption{Conceptual illustration of the motivation for low-light bridge damage detection.}
    \label{fig:introduction}
\end{figure}

\section{Introduction}
\label{sec:introduction}

Bridges are long-serving assets within transportation infrastructure systems, and their safety and serviceability are continuously affected by traffic loading, environmental exposure, and material aging \cite{WANG2026106741}. Structural deterioration often first manifests as surface damage, including cracking, spalling, corrosion, seepage, and localized material loss \cite{koch2015review}. Although externally visible, these manifestations often indicate the degradation of structural durability, declining service performance, and the accumulation of potential maintenance risks. Timely, accurate, and reliable identification of bridge surface damage is therefore essential for structural condition assessment and maintenance and repair decision-making. Efficiently and reliably acquiring and interpreting damage information under practical engineering conditions has consequently become a critical issue in intelligent bridge maintenance and automated inspection \cite{PANIGATI2025106101}.

Against this background, UAV-assisted visual inspection has emerged as an important approach to field data acquisition and damage recognition for bridges \cite{agnisarman2019survey}. Existing studies have extensively investigated UAV inspection path planning and field data acquisition \cite{duque2018synthesis}, with subsequent efforts extending to image-based damage quantification \cite{ellenberg2015use}, deep-learning-based bridge crack recognition \cite{yeum2015vision,cha2017deep}, and the detection of multiple types of surface damage \cite{cha2018autonomous}. These advances have substantially improved the accessibility of bridge undersides, piers, elevated components, and bridges located in complex terrain, while promoting a transition from manual field observation toward an integrated workflow of ``UAV image acquisition--automated damage recognition--condition information extraction'' \cite{seo2018drone}. However, despite the improvements in inspection accessibility and data-acquisition efficiency, the quality of UAV imagery remains susceptible to field imaging conditions. In particular, uncontrollable illumination beneath bridges, within structural interiors, and in locally occluded regions can cause pronounced low-light degradation in the acquired images, thereby compromising subsequent damage information extraction.

More specifically, illumination conditions during bridge inspection are generally difficult to control. Bridge undersides, inter-girder regions, areas near bearings and expansion joints, box-girder interiors, and component surfaces occluded by structural elements are particularly susceptible to shadows, localized underexposure, and substantial brightness variations. Image exposure is also jointly affected by weather conditions, inspection time, and local occlusions. Because these factors are highly site-dependent, it is difficult to consistently acquire images with uniform brightness, sufficient contrast, and clearly visible damage details across different structural components and viewpoints. Low-light conditions further cause underexposure, contrast reduction, amplified noise in dark regions, and blurred damage boundaries, thereby reducing the visibility of small and low-contrast damage. Consequently, illumination variation affects not only the visual quality of UAV imagery but also the ability of detection and segmentation models to recognize, localize, and delineate damaged regions.

Beyond low-light degradation and the associated difficulty of restoring information for multiple damage categories, the development of low-light image restoration and damage recognition models is further constrained by the limited availability of suitable training data. Supervised low-light image restoration ideally requires normal-light--low-light image pairs depicting the same damage locations from the same viewpoints with strict spatial correspondence. During UAV-based bridge inspection, however, flight pose, imaging distance, shadow distribution, weather conditions, and camera parameters are difficult to maintain consistently, making strictly paired field data difficult to acquire. Even when the objective is limited to training low-light damage detectors, a large number of field images covering different bridge types, structural components, damage categories, and illumination conditions must still be collected and precisely annotated. This process is costly and time-consuming, and annotation quality can be adversely affected by severe underexposure and ambiguous damage boundaries. Consequently, constructing controllable training samples in the absence of real paired low-light data, while recovering visual information beneficial to downstream damage recognition, remains a practical challenge for automated bridge inspection.

To address these challenges, this paper presents a task-oriented low-light visual information restoration framework for bridge damage recognition, rather than a generic low-light enhancement method aimed primarily at improving perceptual appearance. At the data level, a physics-grounded ISP-aware synthesis pipeline is used to transform normal-light bridge imagery into spatially aligned low-light samples while retaining the original multi-class object detection and instance segmentation annotations. This provides a basis for supervised restoration training and controlled evaluation of downstream recognition under low illumination. The related synthesis background and detailed implementation are presented in Sections~\ref{sec:related_work} and~\ref{sec:low_light_synthesis}, respectively. At the model level, the Degradation-aware Low-light Mixture-of-Experts Network (DaL-MoE) is proposed to combine degradation-aware guidance estimation, adaptive expert collaboration, and multi-scale feature restoration, thereby balancing noise suppression, illumination regulation, and preservation of damage-relevant structures. The detailed architecture is described in Section~\ref{sec:methodology}. DaL-MoE is trained independently of downstream recognition models and produces standardized RGB images. It can therefore be deployed as a plug-and-play, detector-agnostic restoration front end for different object detection and instance segmentation frameworks. The effectiveness of the framework is evaluated at the levels of low-level restoration quality, downstream damage recognition, and transfer to real UAV inspection imagery.

The main contributions of this paper are summarized as follows:
\begin{enumerate}
    \item To address the scarcity of paired low-light bridge damage data, this paper constructs an ISP-aware dataset comprising normal-light UAV images, spatially aligned synthetic low-light counterparts, and shared annotations for object detection and instance segmentation. Exposure attenuation and sensor noise are simulated in a RAW-like domain while preserving the original damage locations and scene geometry, enabling paired image restoration training and controlled downstream damage recognition evaluation.

    \item Building on the constructed dataset, we propose the Degradation-aware Low-light Mixture-of-Experts Network (DaL-MoE), which integrates degradation-aware guidance estimation, hierarchical multi-scale restoration, and adaptive fusion of parallel experts specialized in noise suppression, color adjustment, and structural-detail recovery. Trained independently as an image-to-image restoration model, DaL-MoE functions as a plug-and-play, detector-agnostic front end that requires neither modification of nor joint optimization with downstream detection or segmentation models.

    \item Extensive experiments provide a multi-level evaluation of full-reference image restoration quality on paired synthetic data, object detection and instance segmentation performance across multiple downstream frameworks, and real-scene transfer using UAV inspection imagery without paired normal-light references. The results demonstrate consistent improvements on the synthetic benchmark and provide both qualitative and quantitative evidence of sim-to-real transfer to practical bridge inspection scenes.
\end{enumerate}

\section{Related Work}
\label{sec:related_work}

To address the pronounced performance degradation of bridge damage recognition under uncontrollable illumination, prior investigations have largely developed along two complementary trajectories: benchmark dataset construction and low-light visual restoration. At the data level, recent efforts focus on expanding defect coverage from crack-only patterns to multi-class surface deteriorations and synthesizing controllable low-light degradation when real paired observations are unavailable. At the methodological level, the objective has evolved from generic global illumination enhancement toward task-oriented restoration that suppresses sensor noise while preserving subtle textures and fine boundaries vital for damage delineation. Accordingly, this section reviews existing bridge damage datasets, generic low-light benchmarks, and degradation synthesis pipelines, followed by an analysis of low-light enhancement and task-oriented adaptive restoration models. Finally, the remaining research gaps across data, architecture, and evaluation paradigms are summarized to establish the positioning of this paper.

Table~\ref{tab:related_work_positioning} systematically summarizes representative research streams across the data and model levels, highlighting the positioning of this paper.

\begin{table*}[pos=h]
\centering
\caption{Positioning of the present study relative to representative data- and model-level research streams for low-light bridge damage analysis.}
\label{tab:related_work_positioning}

\fontsize{6.2}{7.1}\selectfont
\setlength{\tabcolsep}{2.8pt}
\renewcommand{\arraystretch}{1.14}
\setlength{\aboverulesep}{0.35ex}
\setlength{\belowrulesep}{0.35ex}

\begin{tabularx}{\linewidth}{
@{}
>{\RaggedRight\arraybackslash}p{0.075\linewidth}
>{\RaggedRight\arraybackslash}p{0.135\linewidth}
>{\RaggedRight\arraybackslash}p{0.215\linewidth}
>{\RaggedRight\arraybackslash}p{0.215\linewidth}
>{\RaggedRight\arraybackslash}X
@{}}
\toprule
\makecell[c]{\textbf{Research}\\[-1pt]\textbf{level}}
&
\makecell[c]{\textbf{Research}\\[-1pt]\textbf{stream}}
&
\makecell[c]{\textbf{Typical data}\\[-1pt]\textbf{or methods}}
&
\makecell[c]{\textbf{Representative}\\[-1pt]\textbf{capability}}
&
\makecell[c]{\textbf{Main limitation relative}\\[-1pt]\textbf{to the present study}}
\\
\midrule

\multirow{4}{*}{\makecell[l]{Data\\level}}
&
Bridge damage datasets
&
Crack classification, multi-label damage classification, multi-class detection, and segmentation
\cite{dorafshan2018sdnet,mundt2019codebrim,huang2024bridgedamage,li2025gyudet}
&
Provide bridge-specific damage categories, inspection scenes, and spatial annotations at diverse granularities
&
Lack strictly pixel-aligned normal-light--low-light pairs and shared detection/segmentation annotations across both illumination conditions
\\

\arrayrulecolor{black!22}\cmidrule(lr){2-5}\arrayrulecolor{black}

&
Generic low-light datasets
&
Paired low-light restoration, nighttime object detection, and visible--infrared multimodal perception
\cite{wei2018deep,chen2018learning,loh2019exdark,jia2021llvip}
&
Support full-reference restoration or high-level visual perception under low illumination
&
Focus on natural, traffic, and surveillance domains, lacking civil engineering semantics and bridge defect contexts
\\

\arrayrulecolor{black!22}\cmidrule(lr){2-5}\arrayrulecolor{black}

&
Synthetic low-light data
&
sRGB-domain degradation, RAW-domain exposure/noise simulation, and day-to-night synthesis
\cite{yang2020fidelity,brooks2019unprocessing,punnappurath2022daytonight}
&
Provide controllable physical parameters and spatially aligned normal-light--low-light samples
&
Are not tailored to multiple bridge defect categories and lack mechanisms to preserve and evaluate instance-level annotations
\\

\arrayrulecolor{black!22}\cmidrule(lr){2-5}\arrayrulecolor{black}

&
Low-light infrastructure datasets
&
Normal- and low-light crack images and multi-darkness crack segmentation benchmarks
\cite{yao2024cracknex,ren2026lowlightcrack}
&
Provide specialized data for low-light enhancement and crack segmentation in civil structures
&
Remain strictly crack-centric and do not support simultaneous object detection and instance segmentation across multiple defect categories
\\

\midrule

\multirow{3}{*}{\makecell[l]{Model\\level}}
&
Generic low-light enhancement and restoration
&
Retinex decomposition, illumination estimation, deep restoration backbones, and compound degradation recovery
\cite{jobson1997multiscale,guo2017lime,guo2020zero,zamir2020mirnet,Zamir2021Restormer,xu2022snr,retinexformer,yan2025hvi,Feijoo_2025_CVPR}
&
Recover illumination, suppress noise, and restore color and fine details
&
Are optimized and evaluated on pixel reconstruction or perceptual quality, without explicitly preserving defect-critical boundaries and textures
\\

\arrayrulecolor{black!22}\cmidrule(lr){2-5}\arrayrulecolor{black}

&
Low-light infrastructure perception
&
Low-light crack enhancement, reflectance modeling, and enhancement-assisted crack segmentation
\cite{yao2024cracknex,guo2024crackrestoration,ren2026lowlightcrack}
&
Demonstrate the efficacy of illumination restoration for civil infrastructure crack identification
&
Target a single crack category with backbone-coupled pipelines, lacking multi-class evaluation and cross-framework adaptability
\\

\arrayrulecolor{black!22}\cmidrule(lr){2-5}\arrayrulecolor{black}

&
Adaptive and degradation-aware restoration
&
Degradation prompts, conditional restoration, dynamic routing, and mixture-of-experts architectures
\cite{potlapalli2023promptir,jacobs1991adaptive}
&
Adapt restoration behavior dynamically based on input degradation characteristics
&
Target generic restoration, with limited exploration of defect-tailored expert specialization, degradation guidance, and detector-agnostic deployment
\\

\midrule
\rowcolor{oursbg}
\textbf{This paper}
&
Low-light multi-class bridge damage restoration and recognition
&
ISP-aware paired low-light synthesis, degradation-aware MoE restoration, and multi-framework evaluation
&
Bridges physics-grounded synthesis, multi-class visual information recovery, object detection, and instance segmentation
&
Addresses the underexplored intersection of bridge-specific paired data construction, degradation-adaptive restoration, and cross-framework defect recognition
\\

\bottomrule
\end{tabularx}
\end{table*}

\subsection{Datasets for Bridge Damage and Low-Light Vision}
\label{subsec:related_datasets}

Datasets for bridge and concrete surface damage recognition have evolved from binary crack classification toward multi-class damage detection and segmentation. Early benchmarks were predominantly designed for crack categorization. For instance, SDNET2018 collected local image patches from concrete bridge decks, walls, and pavements for binary classification \cite{dorafshan2018sdnet}, while the COncrete DEfect BRidge IMage dataset (CODEBRIM) extended this setting to multi-label classification covering cracks, spalling, efflorescence, exposed reinforcement, and corrosion stains \cite{mundt2019codebrim}. Although these datasets laid the foundation for learning defect-sensitive visual representations, their image-level formulation offers limited support for evaluating spatial localization, boundary delineation, and instance-level geometric properties.

Driven by the need for fine-grained structural condition assessment, recent datasets have substantially enhanced both annotation granularity and damage diversity. The BridgeDamage dataset provides over 2,800 inspection images covering five major surface defect categories for multi-class detection and segmentation \cite{huang2024bridgedamage}. GYU-DET further contributes 11,123 high-resolution images with bounding-box annotations across six defect categories, capturing diverse field environments and natural illumination variations \cite{li2025gyudet}. These resources have expanded the data scale and structural complexity available for automated bridge inspection.

Nevertheless, natural illumination diversity in field imagery is fundamentally distinct from strictly paired normal-light--low-light observations. Existing bridge inspection datasets are primarily tailored for defect perception under standard lighting; they lack pixel-aligned image pairs depicting identical damage instances and scene geometry under both illumination conditions. Furthermore, they do not provide detection and instance segmentation annotations that are shared and verified across paired lighting states. Consequently, these datasets cannot directly supervise low-light image restoration models, nor can they support controlled benchmark analyses to quantitatively isolate the adverse impact of low illumination on downstream defect recognition while holding spatial geometry constant. This limitation is directly related to the difficulty of acquiring strictly paired field observations under practical UAV inspection conditions.

In parallel, the low-light vision community has developed specialized benchmarks for image restoration and high-level visual perception. For low-level restoration, the LOL dataset provides real-world paired low-light and normal-light images \cite{wei2018deep}, while See-in-the-Dark (SID) pairs short-exposure RAW observations with long-exposure reference images for extreme low-light recovery \cite{chen2018learning}. For high-level perception, ExDark establishes a multi-class object detection benchmark under low illumination \cite{loh2019exdark}, and LLVIP contributes spatially aligned visible--infrared image pairs for low-light pedestrian perception \cite{jia2021llvip}. While effective in their intended domains, these benchmarks predominantly feature natural landscapes, urban traffic, and surveillance scenes, exhibiting a substantial domain gap from bridge surface defects and civil engineering semantics. They therefore do not directly satisfy the data requirements of low-light bridge damage restoration and multi-class recognition.

To circumvent the difficulty of collecting real paired data, physics-based low-light synthesis offers a controllable alternative for generating spatially aligned training samples. LOL-v2 demonstrated the efficacy of synthetic low-light subsets for algorithm validation \cite{yang2020fidelity}. Unprocessing pipelines invert the camera image signal processing (ISP) pipeline into a RAW-like domain, realistically simulating exposure attenuation, signal-dependent shot noise, and signal-independent read noise \cite{brooks2019unprocessing}. Similar principles have been applied to synthesize paired daytime and nighttime images for training neural ISPs \cite{punnappurath2022daytonight}. However, existing synthesis frameworks are designed for generic natural imagery rather than civil structures, lacking mechanisms to preserve and reuse fine-grained bounding-box and instance-mask annotations across degraded and restored counterparts. This limitation restricts their direct application to multi-class bridge damage restoration and downstream recognition evaluation.

Currently, low-light datasets dedicated to civil infrastructure remain scarce and strictly crack-centric. CrackNex introduced the LCSD dataset for few-shot crack segmentation under varying illumination \cite{yao2024cracknex}, and GalleryLL-1240 provided crack images across distinct darkness levels for underlit dam galleries \cite{ren2026lowlightcrack}. Nonetheless, actual bridge visual inspection encounters a heterogeneous spectrum of compound deterioration; beyond linear cracks, it involves spalling, seepage, honeycombing, exposed reinforcement, reinforcement corrosion, rust staining, and efflorescence. These defect categories exhibit distinct scale distributions, color contrasts, texture roughness, and topological boundaries, imposing conflicting demands on restoration and feature preservation. Therefore, a low-light bridge dataset that simultaneously provides strict spatial pairing, physics-grounded degradation simulation, and multi-class annotations for both detection and instance segmentation remains an unresolved gap in the literature.

\subsection{Low-Light Image Enhancement and Task-Oriented Restoration}
\label{subsec:low_light_restoration}

Low-light image processing has evolved from conventional heuristic illumination adjustment and Retinex decomposition toward data-driven restoration of compound degradations. Classical Retinex models formulate a low-light observation as the product of reflectance and illumination layers, enhancing visibility by estimating either component \cite{jobson1997multiscale}. Illumination-map estimation methods such as LIME further recover dark-region content via spatially varying illumination estimation paired with structure-preserving constraints \cite{guo2017lime}. Although these approaches established the theoretical foundation for illumination modeling, their performance in complex real-world environments is often degraded by sensor noise amplification, severe color casts, and non-uniform lighting.

With the advent of deep learning, illumination compensation, noise suppression, and detail reconstruction can be jointly optimized. Zero-DCE frames low-light enhancement as a zero-reference pixel-wise curve estimation task, eliminating the dependency on paired training data \cite{guo2020zero}. MIRNet preserves complementary representations across spatial scales through recursive multi-scale feature interaction \cite{zamir2020mirnet}, while Restormer introduces an efficient transposed Transformer architecture to capture long-range token dependencies in high-resolution image restoration \cite{Zamir2021Restormer}. These methodologies have advanced low-light processing from empirical intensity stretching to deep, multi-scale feature modeling.

Recent studies increasingly recognize that low-light degradation is rarely a simple global luminance reduction, but rather an entangled compound process involving severe noise, contrast loss, chromatic distortion, and structural degradation. SNR-Aware enhancement introduces spatial signal-to-noise-ratio priors to dynamically balance local convolutional filtering and non-local feature interactions across heterogeneous degradation zones \cite{xu2022snr}. Retinexformer explicitly estimates illumination cues to guide non-local self-attention and corruption recovery \cite{retinexformer}. HVI-CIDNet designs a decoupled color space to alleviate chromatic distortion and refine photometric mapping under complex lighting \cite{yan2025hvi}, whereas DarkIR systematically addresses the co-occurrence of insufficient illumination, noise, and motion blur in real-world nighttime scenarios \cite{Feijoo_2025_CVPR}. Collectively, these advances highlight a paradigm shift toward the coordinated restoration of multi-faceted physical degradations.

Despite these achievements, prevailing low-light restoration models remain predominantly optimized and benchmarked for pixel-level reconstruction fidelity (e.g., PSNR and SSIM) or human perceptual appearance. However, improvements in visual brightness or generic fidelity metrics do not necessarily guarantee the preservation of weak textures, subtle boundaries, and low-contrast morphological cues indispensable to downstream recognition models. In short, a visually pleasing or brightened image does not inherently translate into superior defect localization and boundary delineation.

This metric discrepancy is particularly pronounced in bridge surface damage inspection. Overly aggressive denoising tends to obliterate thin cracks and subtle spalling edges; unconstrained brightness amplification frequently exaggerates concrete surface roughness, stains, and environmental background clutter; and excessive high-frequency sharpening can introduce pseudo-structural artifacts that misdirect downstream detectors. Consequently, low-light processing for civil inspection must move beyond cosmetic visual enhancement to achieve a rigorous balance among illumination recovery, artifact suppression, and structural preservation, with its utility quantitatively verified on downstream tasks such as object detection and instance segmentation. This requirement is consistent with the task-oriented low-light visual information restoration objective introduced in the Introduction.

A few recent efforts have begun integrating low-light restoration with civil infrastructure inspection. CrackNex derives illumination-invariant reflectance features via Retinex decomposition for few-shot crack segmentation under weak illumination \cite{yao2024cracknex}. Related work has utilized conditional generative models to recover concrete crack images degraded by low light, overexposure, and blur prior to automated assessment \cite{guo2024crackrestoration}. For underlit dam galleries, LL-CrackSeg integrates unsupervised enhancement with a customized segmentation network to improve crack extraction accuracy \cite{ren2026lowlightcrack}. Nevertheless, these methods remain strictly crack-centric, and their enhancement modules are typically tightly coupled with specific segmentation backbones. Unlike relatively regular, linear crack topologies, non-crack bridge deteriorations (such as cavity depths in spalling, diffuse scattering in seepage, or chromatic shifts in corrosion and efflorescence) display much stronger heterogeneity in roughness, local contrast, and region topology. These diverse defects impose conflicting requirements on illumination compensation and structural preservation. Crack-specific strategies cannot be directly generalized to multi-class scenarios, and whether restoration front ends can stably benefit diverse detection and instance segmentation architectures remains insufficiently verified.

Furthermore, the spatial heterogeneity of compound low-light degradation imposes strict demands on the adaptive capacity of restoration networks. A single monolithic pathway often faces conflicting optimization objectives when simultaneously performing denoising, brightness lifting, color correction, and detail enhancement. PromptIR introduces learnable degradation prompts to conditionally modulate network representations across diverse corruption types \cite{potlapalli2023promptir}. The Mixture-of-Experts (MoE) paradigm provides an alternative adaptive framework by allocating specialized sub-networks to learn complementary restoration behaviors, dynamically arbitrated by a lightweight gating network \cite{jacobs1991adaptive}. This mechanism is well-suited for low-light bridge imagery, where the severity of noise, underexposure, and boundary degradation varies across structural elements and spatial regions. Nonetheless, explicitly designing functional experts tailored to bridge defect characteristics, guiding their feature recovery via estimated degradation cues, and deploying the model as a detector-agnostic front end for cross-framework evaluation remain largely unexplored in multi-class bridge inspection.

\subsection{Research Gap and Positioning of This Paper}
\label{subsec:research_gap}

In summary, existing low-light vision research remains misaligned with the practical requirements of UAV-based bridge damage inspection. Although existing bridge damage datasets provide domain-specific annotations for multiple damage categories, they generally lack strictly spatially aligned normal-light--low-light image pairs. By contrast, generic low-light datasets and image restoration methods do not adequately reflect the domain semantics of bridge damage. Existing infrastructure-oriented studies also remain largely focused on cracks as a single damage category or are tightly coupled with specific perception backbones. Therefore, the unresolved issue is not merely how to increase the brightness of low-light bridge images, but how to recover damage-relevant visual information under spatially heterogeneous and compound low-light degradation, such that the restored results can support multi-class damage localization and contour delineation across different recognition frameworks.

Based on the research gaps identified above, this paper seeks to answer the following two questions:
\begin{enumerate}
    \item In the absence of real paired normal-light--low-light UAV bridge imagery, how can controllable training and evaluation conditions be established?

    \item How can a detector-agnostic image restoration model recover visual cues relevant to damage recognition for different object detection and instance segmentation frameworks?
\end{enumerate}

Accordingly, this paper does not focus on low-light enhancement aimed solely at improving perceptual brightness, but rather on recovering the visual evidence required for reliable multi-class bridge damage recognition under adverse illumination conditions.


\section{Methodology}
\label{sec:methodology}

\begin{figure}[pos=h]
    \centering
    \includegraphics[width=\linewidth]{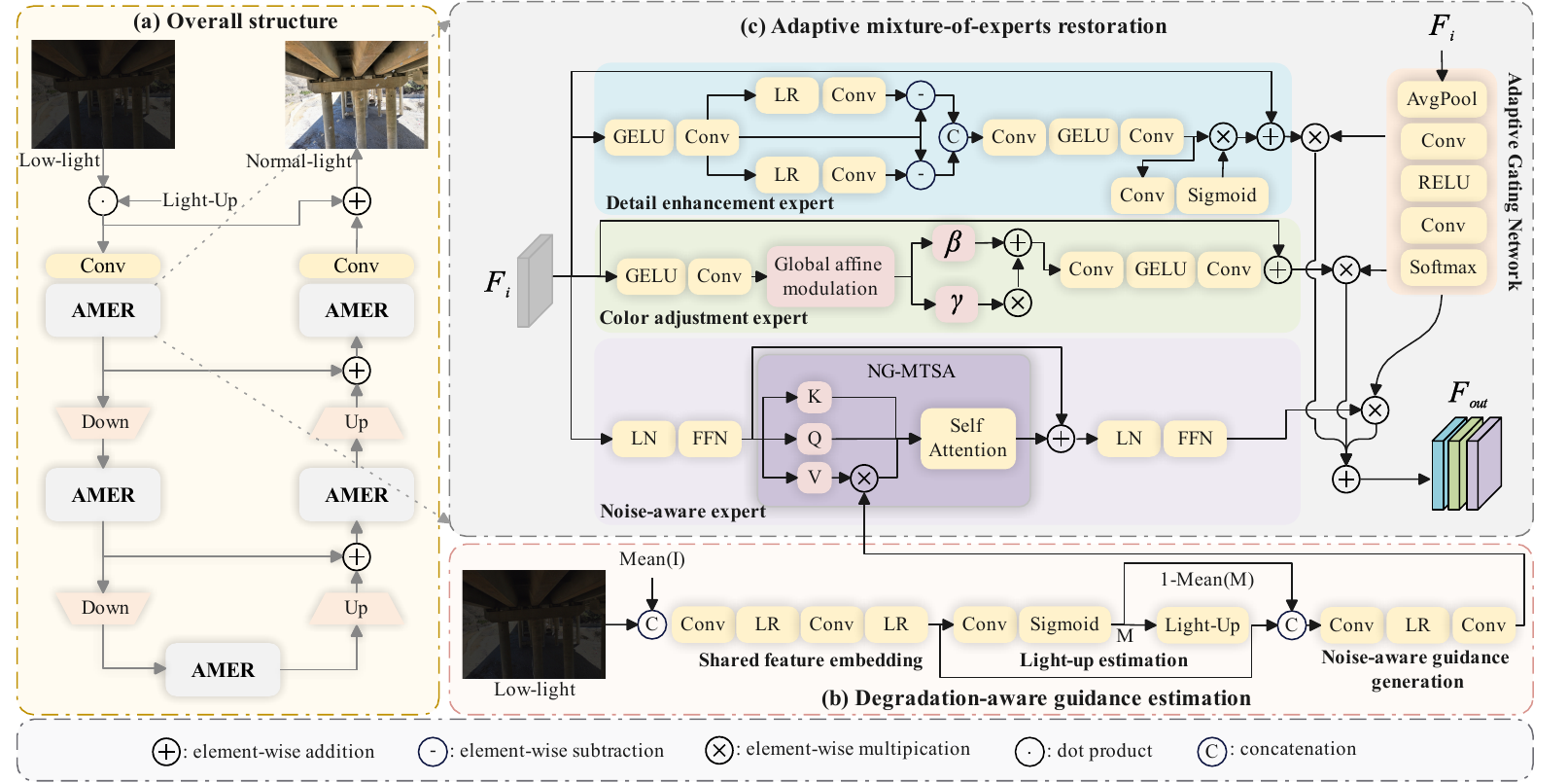}
    \caption{Overview of the proposed DaL-MoE framework.
    (a) Overall architecture, consisting of a Degradation-Aware Guidance Estimation (DAGE) module and a U-shaped Adaptive Mixture-of-Experts Restoration (AMER) network.
    (b) Degradation-Aware Guidance Estimation module, which estimates a spatially varying illumination map and generates noise-aware guidance features from the low-light input.
    (c) Adaptive Mixture-of-Experts Restoration module, where noise-aware, color adjustment, and detail enhancement experts are dynamically integrated by an adaptive gating network for feature restoration.}
    \label{fig:method}
\end{figure}

This section presents the Degradation-aware Low-light Mixture-of-Experts Network (DaL-MoE)
for task-oriented bridge image enhancement under poor illumination. In computer-aided bridge
inspection, low-light degradation is rarely a simple reduction in overall image brightness.
It typically manifests as a compound of spatially coupled phenomena: dark-region noise
amplification, loss of local contrast, tonal and chromatic instability, and weakened
structural boundaries. These degradation factors can collectively obscure damage-relevant
evidence, including crack discontinuities, spalling contours, corrosion traces, seepage
regions, exposed rebar edges, and subtle material texture variations. Accordingly, an
enhancement model intended to support bridge damage analysis must go beyond global brightness
lifting: it must suppress low-light artifacts while preserving the structural cues that are
informative for subsequent detection and instance segmentation.

The term \emph{task-oriented} in this work indicates that the enhancement model is designed
and evaluated with downstream bridge damage analysis as the target application, while the
enhancement network itself remains a detector-agnostic restoration front-end. DaL-MoE
therefore does not require a detection loss or an embedded detector during training. Instead,
its architecture is organized around the visual degradation factors that most directly impair
bridge damage recognition under poor illumination.

The design follows a two-stage principle. In the first stage, the network estimates
illumination- and noise-related guidance from the low-light input image, identifying
spatial regions where degradation is likely to be more severe. In the second stage, a
U-shaped restoration backbone employs an adaptive mixture-of-experts mechanism to balance
complementary restoration behaviors, since noise suppression, brightness correction, and
structural detail preservation may be needed to different degrees across bridge scenes,
viewpoints, and illumination conditions. This principle leads to the two-module architecture illustrated in Fig.~\ref{fig:method}, consisting of a Degradation-Aware Guidance Estimation (DAGE) module and an Adaptive Mixture-of-Experts Restoration (AMER) module.

\textbf{Problem statement.} Given a low-light bridge image $I \in \mathbb{R}^{H \times W \times 3}$, the goal is to
learn an enhancement mapping $\mathcal{N}(\cdot)$ that produces an enhanced image
$\hat{I} \in \mathbb{R}^{H \times W \times 3}$: $\hat{I} = \mathcal{N}(I).$
The enhanced output is expected to exhibit improved visibility, reduced low-light
degradation, and better-preserved damage-relevant structural cues for downstream bridge
damage analysis.

\textbf{Overview.} The overall architecture of DaL-MoE, illustrated in Fig.~\ref{fig:method}(a), consists of two main components: a \textit{Degradation-Aware Guidance Estimation} (DAGE) Module and an \textit{Adaptive Mixture-of-Experts Restoration} (AMER) Module. Given a low-light input image $I$, the degradation-aware guidance estimation module first predicts an illumination map $M$ and extracts a noise-aware guidance feature $F_g$. The illumination map is used to preliminarily compensate for insufficient illumination through residual illumination scaling, producing a lightened image $I_0$, while $F_g$ encodes degradation-related information for subsequent restoration. The lightened image $I_0$ is first encoded into the feature space through a $3\times3$ convolution, and the resulting image feature, together with the guidance feature $F_g$, is then fed into the adaptive mixture-of-experts restoration (AMER) module. Within its U-shaped backbone, $F_g$ is resampled to match the spatial resolution at each scale, while adaptive MoE blocks progressively restore the image representation by dynamically integrating complementary noise-suppression, brightness-correction, and structural-detail recovery behaviors. Finally, the decoded feature is projected into the RGB space through a $3\times3$ convolution and added to $I_0$ via residual reconstruction to obtain the enhanced image $\hat{I}$. In the following, we describe the DAGE module and the AMER module in detail.

\subsection{Degradation-Aware Guidance Estimation Module}
\label{subsec:guidance}

As illustrated in Fig.~\ref{fig:method}(b), the degradation-aware guidance estimation module is designed to characterize spatially varying low-light degradation and provide explicit guidance for subsequent feature restoration. The module consists of three functional components: a \textit{shared feature embedding} block, a \textit{light-up estimation} block, and a \textit{noise-aware guidance generation} block. Given a low-light image $I\in\mathbb{R}^{H\times W\times 3}$, the shared feature embedding block first augments the RGB observation with an illumination-sensitive intensity cue and extracts a compact representation $F_{\mathrm{e}}$. Based on this shared representation, the light-up estimation block predicts a spatially varying map $M$ and produces a preliminarily compensated image $I_0$ through residual illumination scaling. Meanwhile, a noise-related degradation prior $P_n$ is derived from $M$ and combined with $F_{\mathrm{e}}$ in the noise-aware guidance generation block to obtain the guidance feature $F_g$. The overall mapping of the module is expressed as
\begin{equation}
    (F_g,\,M)=\mathcal{P}(I),
    \label{eq:dalmoe_guidance}
\end{equation}
where $M\in[0,1]^{H\times W\times 3}$ denotes the predicted light-up map and $F_g\in\mathbb{R}^{H\times W\times C}$ denotes the noise-aware guidance feature. Consequently, the module provides two complementary outputs for subsequent restoration: the preliminarily compensated image $I_0$ as the restoration input and the degradation-aware feature $F_g$ as conditional guidance.

\paragraph{Shared Feature Embedding Block.}
Directly estimating degradation characteristics from the RGB input alone may provide insufficient illumination context, particularly in spatially non-uniform low-light regions. To introduce an explicit intensity cue while retaining the original chromatic information, the input image is augmented with its channel-wise mean map:
\begin{equation}
    I_{\mathrm{cat}}
    =
    \mathrm{Concat}\!\left(
        I,\,
        \mathrm{Mean}_c(I)
    \right)
    \in
    \mathbb{R}^{H\times W\times 4},
    \label{eq:ngab_concat}
\end{equation}
where $\mathrm{Mean}_c(\cdot)$ denotes averaging along the channel dimension. The resulting four-channel representation jointly preserves the RGB appearance and provides a compact description of the local intensity distribution.

The augmented input is subsequently projected into a $C$-dimensional feature space using a $1\times1$ convolution, followed by a $5\times5$ depthwise convolution for local spatial aggregation:
\begin{equation}
    F_{\mathrm{e}}
    =
    \eta\!\left(
        \mathcal{D}_{5\times5}^{\mathrm{ctx}}
        \!\left(
            \eta\!\left(
                \mathcal{C}_{1\times1}^{\mathrm{emb}}
                (I_{\mathrm{cat}})
            \right)
        \right)
    \right),
    \label{eq:ngab_base_feature}
\end{equation}
where $\eta(\cdot)$ denotes the LeakyReLU activation. The $1\times1$ convolution performs cross-channel feature mixing, whereas the $5\times5$ depthwise convolution aggregates local spatial context with limited computational overhead. The resulting embedding
$F_{\mathrm{e}}\in\mathbb{R}^{H\times W\times C}$ serves as a shared representation for both illumination compensation and degradation-aware guidance generation.

\paragraph{Light-Up Estimation Block.}
Based on the shared feature $F_{\mathrm{e}}$, the light-up estimation block predicts a spatially varying map through a $1\times1$ convolution followed by Sigmoid normalization:
\begin{equation}
    M
    =
    \sigma\!\left(
        \mathcal{C}_{1\times1}^{\mathrm{lum}}
        (F_{\mathrm{e}})
    \right)
    \in
    [0,1]^{H\times W\times 3},
    \label{eq:ngab_lightup_map}
\end{equation}
where $\sigma(\cdot)$ denotes the Sigmoid function. Rather than directly replacing the original image intensity, the estimated map is applied through residual illumination scaling:
\begin{equation}
    I_0
    =
    I\odot(1+M),
    \label{eq:dalmoe_boost}
\end{equation}
where $\odot$ denotes element-wise multiplication. This formulation preserves the original low-light observation while introducing spatially adaptive illumination compensation, thereby providing a better-exposed image representation for subsequent restoration.

\paragraph{Noise-Aware Guidance Generation Block.}
Low-light degradation is generally spatially heterogeneous, with poorly illuminated regions being more susceptible to noise amplification and contrast deterioration. Instead of introducing an additional noise-estimation network or explicit noise supervision, a lightweight degradation prior is derived directly from the predicted map $M$:
\begin{equation}
    P_n
    =
    1-\mathrm{Mean}_c(M)
    \in
    [0,1]^{H\times W\times 1},
    \label{eq:ngab_noise_prior}
\end{equation}
where $P_n$ provides a complementary spatial cue for noise-aware restoration. This complementary representation allows degradation information to be introduced without additional supervisory signals or substantial computational overhead.

The degradation prior $P_n$ is then concatenated with the shared embedding $F_{\mathrm{e}}$ and transformed by two successive $3\times3$ convolutions to generate the final noise-aware guidance feature:
\begin{equation}
    F_g
    =
    \mathcal{C}_{3\times3}^{\mathrm{gout}}
    \!\left(
        \eta\!\left(
            \mathcal{C}_{3\times3}^{\mathrm{gin}}
            \!\left(
                \mathrm{Concat}
                (F_{\mathrm{e}},\,P_n)
            \right)
        \right)
    \right),
    \label{eq:ngab_guidance_feature}
\end{equation}
where $\mathcal{C}_{3\times3}^{\mathrm{gin}}$ and
$\mathcal{C}_{3\times3}^{\mathrm{gout}}$ denote independently parameterized convolutional layers. By jointly modeling the appearance information contained in $F_{\mathrm{e}}$ and the degradation cue encoded by $P_n$, the resulting $F_g$ captures spatially varying information relevant to low-light restoration.

During the subsequent restoration process, $F_g$ is spatially resampled to match the feature resolution at each level of the U-shaped backbone, allowing the restoration blocks to exploit scale-aligned degradation guidance. The preliminarily compensated image $I_0$ and the guidance feature $F_g$ are then jointly forwarded to the Adaptive Mixture-of-Experts Restoration Module, which is introduced in the following subsection.

\subsection{Adaptive Mixture-of-Experts Restoration Module}
\label{subsec:moe_restoration_module}

As illustrated in Fig.~\ref{fig:method}(c), the Adaptive Mixture-of-Experts Restoration Module is designed to adaptively address the coupled degradations present in low-light bridge imagery. At the $i$-th backbone scale, the module receives an image feature $F_i \in \mathbb{R}^{H_i \times W_i \times C_i}$ together with a scale-aligned degradation guidance feature $G_i$. The latter is obtained by spatially aligning the guidance feature $F_g$ generated by the preceding Degradation-Aware Guidance Estimation Module:
\begin{equation}
G_i = \mathcal{C}^{g}_{i}\left(\mathcal{A}_{i}(F_g)\right),
\end{equation}
where $\mathcal{A}_{i}(\cdot)$ denotes spatial resampling to match the resolution of $F_i$, and $\mathcal{C}^{g}_{i}(\cdot)$ denotes a scale-specific $1\times1$ convolution that projects the resampled guidance feature to $C_i$ channels, yielding $G_i \in \mathbb{R}^{H_i\times W_i\times C_i}$.

The module consists of four functional blocks: a \textit{Noise-Aware Expert}, a \textit{Color Adjustment Expert}, a \textit{Detail Enhancement Expert}, and an Adaptive Gating Network. The three expert blocks operate in parallel on the shared input feature $F_i$ and specialize in complementary restoration objectives. Specifically, the \textit{Noise-Aware Expert} additionally incorporates the degradation guidance $G_i$ to suppress noise-related corruption, the \textit{Color Adjustment Expert} regulates illumination-induced chromatic distortions and tonal responses, and the \textit{Detail Enhancement Expert} focuses on recovering local boundaries and structural details. Their outputs are formulated as
\begin{equation}
    F_n = \mathcal{E}_n(F_i,\,G_i), \qquad
    F_b = \mathcal{E}_b(F_i), \qquad
    F_d = \mathcal{E}_d(F_i),
    \label{eq:moe_specialized_experts}
\end{equation}
where $F_n$, $F_c$, and $F_d$ denote the feature representations produced by the \textit{Noise-Aware Expert}, \textit{Color Adjustment Expert}, and \textit{Detail Enhancement Expert}, respectively. Rather than imposing a predefined restoration order on different degradation factors, the parallel expert organization enables each branch to independently learn degradation-specific representations from the same input feature. The \textit{Adaptive Gating Network} subsequently estimates expert-specific routing weights and adaptively integrates the three complementary representations into the restored output feature $F_{\mathrm{out}}$. The four functional blocks are detailed below.

\subsubsection{Noise-Aware Expert}
\label{subsubsec:noise_aware_expert}

The \textit{Noise-Aware Expert} $\mathcal{E}_n$ is designed to suppress noise-related degradations that are particularly pronounced in poorly illuminated regions. Unlike the other two experts, $\mathcal{E}_n$ explicitly incorporates the scale-aligned guidance feature $G_i$, which encodes degradation characteristics estimated from the original low-light input. By introducing such degradation-aware information into feature aggregation, $\mathcal{E}_n$ adaptively suppresses noise-sensitive responses while preserving informative structural content.

Specifically, given the input feature $F_i$ and its corresponding guidance feature $G_i$, $\mathcal{E}_n$ first employs noise-guided multi-head transposed self-attention (NG-MTSA) to perform degradation-aware feature aggregation. The resulting feature is then further refined by a convolutional feed-forward network (FFN):
\begin{equation}
    \widetilde{F}_i =
    F_i +
    \mathrm{NG\text{-}MTSA}(F_i,\,G_i),
    \qquad
    F_n =
    \widetilde{F}_i +
    \mathrm{FFN}\!\left(
        \mathrm{LN}(\widetilde{F}_i)
    \right),
    \label{eq:noise_expert}
\end{equation}
where $\mathrm{LN}(\cdot)$ denotes layer normalization, and $F_n$ is the output of the \textit{Noise-Aware Expert}. The FFN adopts a convolutional architecture consisting of a $1\times1$ convolution, a depthwise $3\times3$ convolution, and a subsequent $1\times1$ convolution, with GELU nonlinearities inserted between consecutive convolutional layers.

We next describe the NG-MTSA operation in detail. Given an input feature
$F \in \mathbb{R}^{H_i \times W_i \times C_i}$,
let $C_i = hd$, where $h$ denotes the number of attention heads and $d$ denotes the feature dimension of each head. We define $N = H_iW_i$ and reshape $F$ into the token representation
$\mathbf{F} \in \mathbb{R}^{N \times C_i}$.
The query, key, and value representations are generated through learnable projections:
\begin{equation}
    Q = \mathbf{F}W_q,
    \qquad
    K = \mathbf{F}W_k,
    \qquad
    V = \mathbf{F}W_v,
    \label{eq:noise_qkv}
\end{equation}
where
$W_q,W_k,W_v \in \mathbb{R}^{C_i \times C_i}$
are learnable projection matrices. The projected representations are partitioned into $h$ attention heads, yielding
$Q_j,K_j,V_j \in \mathbb{R}^{N \times d}$
for the $j$-th head. Similarly, the scale-aligned guidance feature $G_i$ is reshaped and partitioned into
$G_j \in \mathbb{R}^{N \times d}$.

To explicitly incorporate degradation-aware information into feature aggregation, the guidance representation is injected into the value branch through element-wise modulation:
\begin{equation}
    \bar{V}_j = V_j \odot G_j,
    \label{eq:noise_value_modulation}
\end{equation}
where $\odot$ denotes element-wise multiplication. In this manner, the degradation cues encoded in $G_i$ directly regulate the feature responses participating in attention aggregation, allowing the network to attenuate noise-sensitive activations while retaining structurally informative responses.

Instead of constructing spatial token-to-token affinities, whose computational cost grows quadratically with the number of spatial locations, NG-MTSA performs attention in a transposed channel space. Specifically, the query and key representations are first transposed and normalized along the token dimension:
\begin{equation}
    \widehat{Q}_j = \ell_2(Q_j^{\top}),
    \qquad
    \widehat{K}_j = \ell_2(K_j^{\top}),
    \label{eq:noise_l2_norm}
\end{equation}
where
$\widehat{Q}_j,\widehat{K}_j \in \mathbb{R}^{d \times N}$,
and $\ell_2(\cdot)$ denotes $\ell_2$ normalization along the token dimension. The channel-wise attention matrix for the $j$-th head is then computed as
\begin{equation}
    A_j =
    \mathrm{Softmax}\!\left(
        \lambda_j
        \widehat{K}_j
        \widehat{Q}_j^{\top}
    \right)
    \in
    \mathbb{R}^{d \times d},
    \label{eq:noise_attention_matrix}
\end{equation}
where $\lambda_j$ is a learnable rescaling parameter associated with the $j$-th attention head. Consequently, NG-MTSA models global inter-channel dependencies through a $d \times d$ attention matrix without explicitly constructing an $N \times N$ spatial affinity matrix.

The degradation-guided attention output of each head is obtained by aggregating the modulated value representation:
\begin{equation}
    Y_j =
    \left(
        A_j \bar{V}_j^{\top}
    \right)^{\top}
    \in
    \mathbb{R}^{N \times d}.
    \label{eq:noise_attention_output}
\end{equation}

The outputs of all attention heads are subsequently concatenated and projected back to the original feature dimension. Since transposed channel attention mainly captures global inter-channel dependencies, an additional local positional branch is introduced to complement the attention output with local spatial information. In particular, this branch operates on the unmodulated value representation $V$, such that the degradation guidance only affects the global attention aggregation and does not directly alter the local positional cues. The complete NG-MTSA operation is formulated as
\begin{equation}
    \mathrm{NG\text{-}MTSA}(F,\,G_i)
    =
    \mathcal{R}^{-1}\!\left(
        \mathrm{Concat}_{j=1}^{h}(Y_j)W_o
    \right)
    +
    \mathrm{Pos}\!\left(
        \mathcal{R}^{-1}(V)
    \right),
    \label{eq:ng_mtsa_output}
\end{equation}
where
$W_o \in \mathbb{R}^{C_i \times C_i}$
denotes the output projection matrix,
$\mathcal{R}^{-1}(\cdot)$ reshapes the token representation back to its spatial form, and
$\mathrm{Pos}(\cdot)$ denotes the local positional branch implemented by two successive depthwise $3\times3$ convolutions with an intermediate GELU activation.

By combining degradation-guided value modulation, transposed channel-wise attention, and local positional compensation, NG-MTSA enables $\mathcal{E}_n$ to selectively suppress noise-related feature responses while preserving global structural dependencies and local spatial details. The resulting feature $F_n$ serves as the noise-aware representation produced by $\mathcal{E}_n$ for subsequent feature fusion and reconstruction.

\subsubsection{Color Adjustment Expert}
\label{subsubsec:color_adjustment_expert}
The \textit{Color Adjustment Expert} $\mathcal{E}_c$ is designed to correct color distortions and illumination-induced chromatic shifts in low-light features. In poorly illuminated regions, insufficient exposure often leads not only to reduced brightness but also to inaccurate color responses and uneven tonal distributions. To address these degradations, $\mathcal{E}_c$ combines global channel-wise affine modulation with local chromatic refinement, enabling adaptive adjustment of both global color statistics and spatially varying chromatic responses.

\paragraph{Color--Illumination Embedding.}
The shared input feature is first projected into a latent color--illumination representation:
\begin{equation}
    F_c =
    \delta\!\left(
        \mathcal{C}_{1\times1}^{\mathrm{col}}(F_i)
    \right),
    \label{eq:color_embedding}
\end{equation}
where $\delta(\cdot)$ denotes GELU. The $1\times1$ convolution reorganizes cross-channel responses and produces a compact representation that facilitates subsequent color and illumination modulation.

\paragraph{Global Affine Modulation.}
To capture global color and illumination statistics, a channel descriptor is first extracted from $F_c$ through global average pooling. The descriptor is then processed by two successive $1\times1$ convolutions with an intermediate ReLU activation to estimate channel-wise affine parameters:
\begin{equation}
    [\gamma,\,\beta] =
    \mathrm{Split}\!\left(
        \mathcal{C}_{1\times1}^{\mathrm{aff2}}\!\left(
            \rho\!\left(
                \mathcal{C}_{1\times1}^{\mathrm{aff1}}\!\left(
                    \mathrm{GAP}(F_c)
                \right)
            \right)
        \right)
    \right),
    \label{eq:color_affine_params}
\end{equation}
where $\rho(\cdot)$ denotes ReLU,
$\mathcal{C}_{1\times1}^{\mathrm{aff1}}$ reduces the channel dimension to $C_i/4$, and
$\mathcal{C}_{1\times1}^{\mathrm{aff2}}$ expands it to $2C_i$.
$\mathrm{Split}(\cdot)$ evenly partitions the output along the channel dimension, yielding
$\gamma,\beta \in \mathbb{R}^{C_i \times 1 \times 1}$.

The resulting affine parameters are spatially broadcast and applied to $F_c$ through a residual affine transformation:
\begin{equation}
    F_a =
    F_c \odot (1+\gamma) + \beta.
    \label{eq:color_affine}
\end{equation}
The multiplicative term adaptively rescales channel responses associated with color and illumination characteristics, while the additive term compensates for global chromatic and tonal offsets. The residual parameterization $(1+\gamma)$ provides an identity-preserving modulation path, allowing the expert to adjust degraded color responses without unnecessarily perturbing informative feature components.

\paragraph{Local Chromatic Refinement.}
Global channel-wise modulation alone is insufficient to characterize spatially varying color distortions caused by non-uniform illumination. Therefore, $F_a$ is further processed by a local chromatic refinement operator:
\begin{equation}
    F_{\mathrm{out}}^{c} =
    \mathcal{C}_{1\times1}^{\mathrm{cout}}\!\left(
        \delta\!\left(
            \mathcal{D}_{3\times3}^{\mathrm{cloc}}(F_a)
        \right)
    \right)
    + F_i,
    \label{eq:color_output}
\end{equation}
where
$\mathcal{D}_{3\times3}^{\mathrm{cloc}}$ captures local spatial and chromatic variations, and
$\mathcal{C}_{1\times1}^{\mathrm{cout}}$ projects the refined representation back to the original channel dimension. The residual connection from $F_i$ preserves the underlying structural representation while allowing $\mathcal{E}_c$ to selectively correct color distortions and locally varying chromatic responses. The resulting $F_{\mathrm{out}}^{c}$ provides a color-adjusted representation complementary to the outputs of the other expert branches.

\subsubsection{Detail Enhancement Expert}
\label{subsubsec:detail_enhancement_expert}

The \textit{Detail Enhancement Expert} $\mathcal{E}_d$ operates directly on the shared input feature $F_i$ and focuses on recovering local structural information that can be weakened under low-light conditions. Bridge damage often contains fine and low-contrast morphological cues, including thin cracks, spalling boundaries, exposed rebar edges, corrosion contours, and local texture discontinuities. To enhance such structure-sensitive information, the expert combines learnable local difference extraction with spatially adaptive structural refinement.

\paragraph{Structural Feature Projection.}
The input feature $F_i$ is first projected into a compact structural representation with half of the original channel dimension:
\begin{equation}
    F_p =
    \delta\!\left(
        \mathcal{C}_{1\times1}^{\mathrm{proj}}(F_i)
    \right)
    \in
    \mathbb{R}^{H_i \times W_i \times C_i/2},
    \label{eq:detail_projection}
\end{equation}
where $\delta(\cdot)$ denotes GELU. This projection reduces channel redundancy and provides a compact representation for subsequent local structure extraction.

\paragraph{Learnable Difference Extraction.}
Two independent depthwise difference branches are employed to capture complementary local structural variations:
\begin{equation}
    D_x =
    \eta\!\left(
        \mathcal{D}_{3\times3}^{x}(F_p)-F_p
    \right),
    \qquad
    D_y =
    \eta\!\left(
        \mathcal{D}_{3\times3}^{y}(F_p)-F_p
    \right),
    \label{eq:detail_difference}
\end{equation}
where $\eta(\cdot)$ denotes LeakyReLU and
$\mathcal{D}_{3\times3}^{x}$ and
$\mathcal{D}_{3\times3}^{y}$ denote two independently parameterized depthwise convolutional filters. Subtracting the input feature from each convolutional response suppresses slowly varying components and emphasizes local differential responses associated with boundaries and texture variations. Unlike fixed gradient operators, the learnable difference branches can adapt their local response patterns to structural characteristics captured during training.

\paragraph{Structural Reconstruction and Spatial Gating.}
The two differential responses are concatenated along the channel dimension and reconstructed through a $1\times1$--GELU--$3\times3$ transformation:
\begin{equation}
    F_s =
    \mathcal{C}_{3\times3}^{\mathrm{rec}}\!\left(
        \delta\!\left(
            \mathcal{C}_{1\times1}^{\mathrm{fus}}\!\left(
                \mathrm{Concat}(D_x,\,D_y)
            \right)
        \right)
    \right),
    \label{eq:detail_reconstruction}
\end{equation}
where
$\mathcal{C}_{1\times1}^{\mathrm{fus}}$ integrates the complementary difference responses across channels and
$\mathcal{C}_{3\times3}^{\mathrm{rec}}$ further aggregates local spatial context to reconstruct structure-sensitive features.

A spatial gating map is subsequently generated to adaptively emphasize informative structural responses:
\begin{equation}
    A_s =
    \sigma\!\left(
        \mathcal{C}_{3\times3}^{\mathrm{gate}}(F_s)
    \right)
    \in
    [0,1]^{H_i \times W_i \times 1},
    \label{eq:detail_gate}
\end{equation}
where $\sigma(\cdot)$ denotes the Sigmoid function. The single-channel gating map $A_s$ is broadcast across feature channels to modulate the reconstructed structural representation. The final output of the expert is obtained through gated residual fusion:
\begin{equation}
    F_d =
    F_s \odot A_s + F_i.
    \label{eq:detail_output}
\end{equation}
By combining learnable difference extraction, structural reconstruction, and spatially adaptive gating, the \textit{Detail Enhancement Expert} strengthens boundary- and texture-sensitive responses while preserving the underlying feature representation through the residual connection. The resulting $F_d$ constitutes the structure-oriented representation used in adaptive expert fusion.

\subsubsection{Adaptive Gating Network}
\label{subsubsec:gating_network}

The three expert branches focus on different restoration objectives and therefore need not contribute equally under different degradation conditions. The relative importance of noise suppression, illumination correction, and structural-detail recovery can vary across input images and feature scales. A fixed combination of the three expert outputs is therefore insufficient to fully exploit their complementary specialization. The \textit{Adaptive Gating Network} addresses this issue by estimating sample-adaptive routing weights and dynamically integrating the expert representations.

Since $F_n$ incorporates the scale-aligned degradation guidance $G_i$, it is used to construct the routing descriptor. Global average pooling first aggregates its spatial responses into a compact representation:
\begin{equation}
    z =
    \mathrm{GAP}(F_n)
    \in
    \mathbb{R}^{C_i}.
    \label{eq:gating_descriptor}
\end{equation}
The descriptor is subsequently processed by two successive $1\times1$ convolutions with an intermediate ReLU activation to generate the routing logits:
\begin{equation}
    r =
    \mathcal{C}_{1\times1}^{\mathrm{rout}}\!\left(
        \rho\!\left(
            \mathcal{C}_{1\times1}^{\mathrm{rin}}(z)
        \right)
    \right)
    \in
    \mathbb{R}^{3},
    \label{eq:gating_logits}
\end{equation}
where $\rho(\cdot)$ denotes ReLU,
$\mathcal{C}_{1\times1}^{\mathrm{rin}}$ reduces the channel dimension to $C_i/2$, and
$\mathcal{C}_{1\times1}^{\mathrm{rout}}$ projects the compact descriptor to three expert-specific logits.

The routing logits are normalized using Softmax to obtain the expert weights:
\begin{equation}
    [w_n,\,w_b,\,w_d]
    =
    \mathrm{Softmax}(r),
    \label{eq:gating_weights}
\end{equation}
where $w_n$, $w_c$, and $w_d$ correspond to the \textit{Noise-Aware Expert}, \textit{Color Adjustment Expert}, and \textit{Detail Enhancement Expert}, respectively, and satisfy
\begin{equation}
    w_n + w_b + w_d = 1,
    \qquad
    w_n,\,w_b,\,w_d \geq 0.
    \label{eq:gating_sum}
\end{equation}

Finally, the output of the Adaptive Mixture-of-Experts Restoration Module is obtained by weighted fusion of the three parallel expert representations:
\begin{equation}
    F_{\mathrm{out}}
    =
    w_nF_n + w_bF_b + w_dF_d.
    \label{eq:gating_fusion}
\end{equation}
Each scalar routing weight is broadcast across the spatial and channel dimensions of its corresponding expert feature. Through this adaptive fusion mechanism, the module dynamically balances noise suppression, illumination and chromatic adjustment, and structural-detail recovery according to the degradation characteristics of the current input representation, yielding the restored feature $F_{\mathrm{out}}$ for subsequent feature propagation in the restoration network.

\section{Low-light Data Synthesis Pipeline}
\label{sec:low_light_synthesis}

Acquiring strictly paired normal-light and low-light images under identical viewpoints is challenging in real-world scenarios. Therefore, we adopt an ISP-aware degradation pipeline to synthesize low-light observations from normal-light images. Unlike direct intensity attenuation in the sRGB domain, this pipeline first unprocesses a normal-light image into a RAW-like linear representation, introduces low-light degradation in the sensor domain, and then renders the corrupted signal back to the sRGB space. This procedure better approximates the physical image formation process under low illumination.

Given a normal-light image $I^n\in[0,1]^{H\times W\times 3}$, the synthetic low-light image $I^l$ is generated as
\begin{equation}
I^l=\mathcal{P}_{\mathrm{ISP}}\left(
\mathcal{D}_{\mathrm{raw}}\left(
\mathcal{P}_{\mathrm{ISP}}^{-1}(I^n)
\right)
\right),
\label{eq:low_light_synthesis}
\end{equation}
where $\mathcal{P}_{\mathrm{ISP}}^{-1}(\cdot)$ denotes the inverse ISP unprocessing operation, $\mathcal{D}_{\mathrm{raw}}(\cdot)$ represents RAW-domain low-light corruption, and $\mathcal{P}_{\mathrm{ISP}}(\cdot)$ denotes the forward ISP rendering process. 

\textbf{RGB-to-RAW unprocessing.}
The normal-light sRGB image is first mapped back to a RAW-like sensor representation by approximately reversing the camera rendering process. Following the convention of \cite{brooks2019unprocessing}, we adopt a three-channel RAW-like linear representation instead of a single-channel Bayer mosaic, which allows color transformations to operate directly on RGB triples. The unprocessing proceeds by inverting the four standard ISP stages in reverse order:

\textit{(i) Inverse tone mapping.} The nonlinear sRGB signal is first linearized as
\begin{equation}
I_t=\frac{1}{2}-\sin\left(\frac{\sin^{-1}(1-2I^n)}{3}\right),
\label{eq:inverse_tone}
\end{equation}
where $I_t$ denotes the inverse-tone-mapped image.

\textit{(ii) Inverse gamma correction.} The linear RGB signal is then recovered via
\begin{equation}
I_{\mathrm{lin}}=\max(I_t,\epsilon)^{\gamma},
\label{eq:inverse_gamma}
\end{equation}
where $\epsilon=10^{-8}$ is used for numerical stability and $\gamma\sim\mathcal{U}(2.0,3.5)$.

\textit{(iii) Color space conversion (RGB $\rightarrow$ camera RGB).} To account for camera-dependent color responses, a camera color correction matrix is randomly selected from a predefined pool and combined with the standard RGB-to-XYZ matrix to obtain a row-normalized RGB-to-camera matrix $M_{\mathrm{rgb}\rightarrow\mathrm{cam}}$. For each pixel location $p$, the camera-domain image is computed as
\begin{equation}
I_{\mathrm{cam}}(p)=M_{\mathrm{rgb}\rightarrow\mathrm{cam}} \, I_{\mathrm{lin}}(p).
\label{eq:rgb_to_cam}
\end{equation}

\textit{(iv) White balance removal.} Finally, the camera-applied white balance is removed to obtain the RAW-like signal:
\begin{equation}
I_{\mathrm{raw}} = I_{\mathrm{cam}} \odot \mathbf{g}_{\mathrm{wb}}, \quad \mathbf{g}_{\mathrm{wb}} = g_{\mathrm{rgb}} \cdot \left[\frac{1}{g_r},\,1,\,\frac{1}{g_b}\right]^{\!\top},
\label{eq:inverse_wb}
\end{equation}
where $\odot$ denotes element-wise multiplication, $\mathbf{g}_{\mathrm{wb}}\in\mathbb{R}^3$ is the channel-wise gain vector, $g_r\sim\mathcal{U}(1.9,2.4)$, $g_b\sim\mathcal{U}(1.5,1.9)$, and $g_{\mathrm{rgb}}\sim\mathcal{N}(0.8,0.1^2)$.

\textbf{RAW-domain low-light corruption.}
We first attenuate the exposure while introducing spatially varying illumination to simulate non-uniform underexposure caused by structural occlusion and local shadows:
\begin{equation}
I_{\mathrm{dark}} = k \, L \odot I_{\mathrm{raw}},
\label{eq:exposure}
\end{equation}
where the global exposure ratio $k$ is sampled from a truncated normal distribution within $[0.01,0.1]$, with mean $0.1$ and standard deviation $0.08$. Here, $L \in \mathbb{R}^{H\times W\times 1}$ denotes a smooth spatial illumination field that models locally varying exposure conditions. Specifically, a low-resolution random field is first generated, interpolated to the image resolution, and Gaussian-smoothed to produce gradual illumination variations. The resulting field is rescaled to $[0.5,1.5]$ and normalized around unity, such that $k$ controls the overall exposure level while $L$ introduces spatially heterogeneous underexposure. Sensor noise is then injected by combining signal-dependent shot noise and signal-independent read noise:
\begin{equation}
\widetilde{I}_{\mathrm{raw}} = I_{\mathrm{dark}}+n, \quad n\sim\mathcal{N}\left(0,\lambda_s I_{\mathrm{dark}}+\lambda_r\right),
\label{eq:noise}
\end{equation}
where $\lambda_s$ and $\lambda_r$ denote the shot-noise and read-noise levels, respectively. Since both variances must be positive and span several orders of magnitude, they are sampled in the logarithmic domain. The shot-noise level is drawn as
\begin{equation}
u_s \sim \mathcal{U}\left(\log(10^{-4}), \log(1.2\times10^{-2})\right), \quad \lambda_s = \exp(u_s),
\label{eq:shot_noise_sampling}
\end{equation}
and the read-noise level is generated from a log-linear camera noise model conditioned on $u_s$:
\begin{equation}
u_r = 2.18\,u_s + 1.20 + \xi, \quad \xi\sim\mathcal{N}(0,0.26^2), \quad \lambda_r = \exp(u_r).
\label{eq:read_noise_sampling}
\end{equation}
This conditional sampling preserves the empirical correlation between shot and read noise while guaranteeing $\lambda_s>0$ and $\lambda_r>0$.

\textbf{RAW-to-RGB rendering.}
After RAW-domain corruption, the degraded signal is rendered back to the sRGB space through the forward ISP pipeline. Quantization noise is first added to simulate analog-to-digital conversion:
\begin{equation}
I_q = \widetilde{I}_{\mathrm{raw}} + q, \quad q\sim\mathcal{U}\left(-\frac{1}{2^{B+1}},\frac{1}{2^{B+1}}\right),
\label{eq:quantization}
\end{equation}
where $B\in\{12,14,16\}$ denotes the simulated bit depth. The forward ISP $\mathcal{P}_{\mathrm{ISP}}(\cdot)$ then applies the four reciprocal operations of the unprocessing stage in forward order: \textit{(i)~white balance restoration}, \textit{(ii)~color space conversion via $M_{\mathrm{cam}\rightarrow\mathrm{rgb}}=M_{\mathrm{rgb}\rightarrow\mathrm{cam}}^{-1}$}, \textit{(iii)~gamma encoding with exponent $1/\gamma$}, and \textit{(iv)~tone mapping}, followed by clipping to the valid range:
\begin{equation}
I^l = \mathrm{clip}\left(\mathcal{P}_{\mathrm{ISP}}(I_q),\,0,\,1\right).
\label{eq:forward_isp}
\end{equation}
The rendered image $I^l\in[0,1]^{H\times W\times 3}$ serves as the synthetic low-light counterpart of $I^n$ for model training and evaluation.

\section{Experiments and Geometric Evaluation}

This section presents the experimental setup and evaluates the proposed DaL-MoE in terms of image enhancement quality and downstream bridge damage detection and instance segmentation performance. We first describe the experimental platform, datasets, implementation details, and evaluation protocols, and then present quantitative and qualitative comparisons, ablation studies, and real-world evaluations.

\subsection{Experimental platform}
\label{subsec:experimental_platform}

All experiments were conducted on a workstation equipped with an AMD Ryzen 9 9950X central processing unit (CPU), 96 GB of random access memory (RAM), and an NVIDIA GeForce RTX 5090 graphics processing unit (GPU) with 32 GB of video random access memory (VRAM), running the Linux operating system.

\begin{table}[pos=t]
\centering
\caption{Overview and class-wise instance distribution of the UAV bridge defect dataset and paired low-light benchmark.}
\label{tab:dataset_overview}

\begingroup
\fontsize{6.55}{7.55}\selectfont
\setlength{\tabcolsep}{2.5pt}
\renewcommand{\arraystretch}{1.08}
\setlength{\aboverulesep}{0.30ex}
\setlength{\belowrulesep}{0.30ex}
\setlength{\cmidrulesep}{0.12ex}

\begin{threeparttable}

\begin{minipage}{\linewidth}
\raggedright
\textbf{Panel A. Dataset acquisition, partitioning, annotation, and benchmark construction}
\end{minipage}

\vspace{1.4pt}

\begin{tabularx}{\linewidth}{
@{}
>{\RaggedRight\arraybackslash}p{0.29\linewidth}
>{\RaggedRight\arraybackslash}X
@{}}

\toprule
\textbf{Item} & \textbf{Content} \\
\midrule

\rowcolor{summarygray}
\multicolumn{2}{@{}l@{}}{%
    \strut\textbf{Field acquisition}%
}
\\[-0.15ex]

Number of bridges
& 7 \\

Inspection contexts
& 4, including urban viaducts, mountainous bridges, railway bridges, and long-span highway bridges \\

Acquisition system
& UAV equipped with a DJI H30T imaging system \\

Original image resolution
& $4032 \times 3024$ pixels \\

\addlinespace[0.45ex]

\rowcolor{summarygray}
\multicolumn{2}{@{}l@{}}{%
    \strut\textbf{Dataset composition and partitioning}%
}
\\[-0.15ex]

Normal-light source images
& 2,895 \\

Train / test split
& 2,413 / 482 images \\

Annotated defect instances
& 19,582, including 15,731 training instances and 3,851 test instances \\

Defect categories
& Spalling, Crack, Hole, Honeycombing, Seepage, Rebar Corrosion, Rust, and Efflorescence \\

Spatial partitioning
& Group-wise partition using mutually exclusive bridge spans or mileage sections rather than random image-level sampling \\

Leakage control
& Adjacent or overlapping views from the same span or mileage section were assigned to one subset; the split was fixed before low-light synthesis \\

\addlinespace[0.45ex]

\rowcolor{summarygray}
\multicolumn{2}{@{}l@{}}{%
    \strut\textbf{Annotation and paired benchmark construction}%
}
\\[-0.15ex]

Manual annotation source
& Closed polygon contours manually delineated for individual defect instances \\

Derived task annotations
& Axis-aligned bounding boxes computed from polygon-coordinate extrema; instance masks generated by rasterizing and filling the same polygons \\

Paired low-light benchmark
& 2,895 spatially aligned normal-light--low-light pairs generated using the ISP-aware synthesis pipeline, with annotations shared across both illumination conditions \\

Supported evaluations
& Full-reference image enhancement, object detection, and instance segmentation \\

\bottomrule
\end{tabularx}

\vspace{4.0pt}

\begin{minipage}{\linewidth}
\raggedright
\textbf{Panel B. Class-wise instance distribution across the training and test sets}
\end{minipage}

\vspace{1.4pt}

\begin{tabularx}{\linewidth}{
@{}
>{\centering\arraybackslash}p{0.09\linewidth}
>{\RaggedRight\arraybackslash}p{0.22\linewidth}
*{3}{>{\centering\arraybackslash}X}
@{}}

\toprule

\makecell[c]{\textbf{Class}\\[-0.5pt]\textbf{ID}}
&
\textbf{Class name}
&
\makecell[c]{\textbf{Train}\\[-0.5pt]\textbf{(2,413 imgs)}}
&
\makecell[c]{\textbf{Test}\\[-0.5pt]\textbf{(482 imgs)}}
&
\makecell[c]{\textbf{Total}\\[-0.5pt]\textbf{(2,895 imgs)}}
\\

\midrule

0
& Spalling
& 3,568 (22.68\%)
& 809 (21.01\%)
& 4,377 (22.35\%)
\\

1
& Crack
& 1,373 (8.73\%)
& 456 (11.84\%)
& 1,829 (9.34\%)
\\

2
& Hole
& 4,315 (27.43\%)
& 1,043 (27.08\%)
& 5,358 (27.36\%)
\\

3
& Honeycombing
& 484 (3.08\%)
& 185 (4.80\%)
& 669 (3.42\%)
\\

4
& Seepage
& 745 (4.74\%)
& 200 (5.19\%)
& 945 (4.83\%)
\\

5
& Rebar corrosion
& 1,695 (10.77\%)
& 362 (9.40\%)
& 2,057 (10.50\%)
\\

6
& Rust
& 845 (5.37\%)
& 239 (6.21\%)
& 1,084 (5.54\%)
\\

7
& Efflorescence
& 2,706 (17.20\%)
& 557 (14.46\%)
& 3,263 (16.66\%)
\\

\midrule

\rowcolor{summarygray}
\multicolumn{2}{@{}l@{}}{\textbf{Total}}
&
\textbf{15,731 (100.00\%)}
&
\textbf{3,851 (100.00\%)}
&
\textbf{19,582 (100.00\%)}
\\

\bottomrule
\end{tabularx}

\vspace{1.2pt}

\begin{tablenotes}[flushleft]
\fontsize{5.90}{6.85}\selectfont
\item[]
\textit{Note:}
Image counts and split sizes refer to the normal-light source images. One spatially aligned synthetic low-light counterpart was generated for each source image. Training and test images were assigned according to mutually exclusive bridge spans or mileage sections, and adjacent or spatially overlapping UAV views were not shared across subsets. Percentages in Panel B indicate column-wise class proportions.
\end{tablenotes}

\end{threeparttable}
\endgroup

\vspace{-3pt}
\end{table}

\subsection{Datasets}
\label{subsec:datasets}

\begin{figure}[pos=t]
    \centering
    \includegraphics[width=\linewidth]{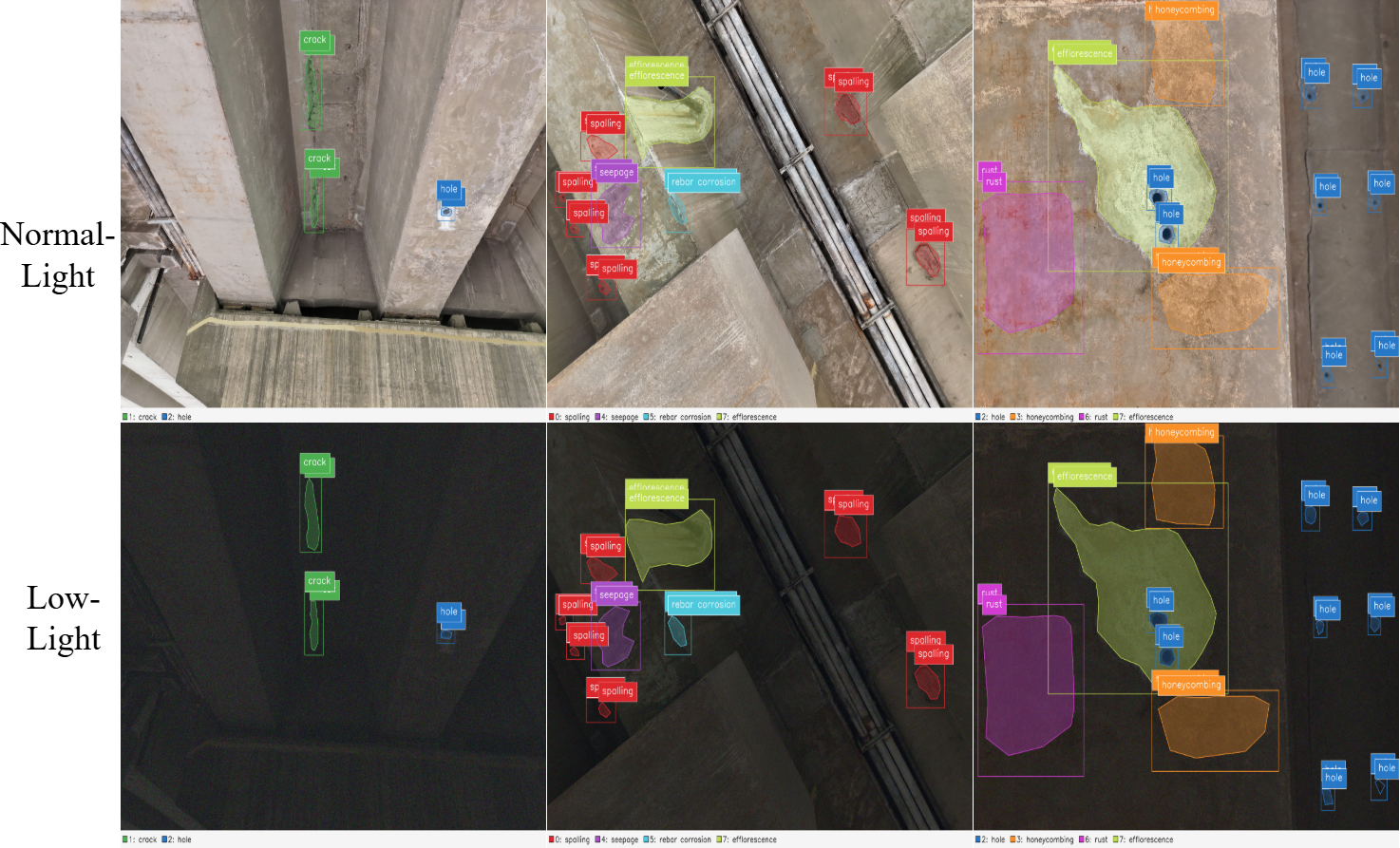}
    \caption{Visualization of defect annotations shared between paired normal-light and synthetic low-light images.}
    \label{fig:paired_annotation_visualization}
\end{figure}

A total of 2,895 images were collected from real-world UAV bridge inspection scenarios covering seven bridges and spatially distinct bridge spans and mileage sections. All images were acquired using a UAV equipped with a DJI H30T imaging system and have an original resolution of $4032 \times 3024$ pixels. To reduce information leakage caused by adjacent viewpoints or repeated local scenes, the dataset was partitioned according to spatially disjoint bridge spans or mileage sections rather than through random image-level sampling. Adjacent or overlapping views from the same span or mileage section were assigned to the same subset. The final dataset contains 2,413 training images and 482 test images.

Eight representative bridge surface defect categories were considered: spalling, crack, hole, honeycombing, seepage, rebar corrosion, rust, and efflorescence. Each defect instance was manually delineated as a closed polygon contour, from which the axis-aligned bounding box and instance segmentation mask were automatically derived. This procedure ensured spatial consistency between the detection and segmentation annotations. As summarized in Table~\ref{tab:datasetorig_distribution}, the dataset contains 19,582 annotated defect instances, including 15,731 instances (80.33\%) in the training set and 3,851 instances (19.67\%) in the test set. Hole is the most frequent category in both subsets, accounting for 27.43\% of the training instances and 27.08\% of the test instances, with an overall proportion of 27.36\%. It is followed by spalling, which accounts for 22.68\% of the training set, 21.01\% of the test set, and 22.35\% of all instances, and efflorescence, which accounts for 17.20\%, 14.46\%, and 16.66\%, respectively. The remaining categories account for 10.50\% (rebar corrosion), 9.34\% (crack), 5.54\% (rust), 4.83\% (seepage), and 3.42\% (honeycombing) of all annotated instances. Thus, the training and test sets exhibit broadly comparable, although not identical, class-wise distributions.

Because strictly paired normal-light and low-light images are difficult to acquire in real-world inspection scenarios, an ISP-aware low-light synthesis pipeline was applied to the normal-light images to generate spatially aligned low-light counterparts. The dataset partition was established before synthesis, such that each source image and its derived low-light image remained in the same subset. Since the synthesis process does not alter the damage locations or scene geometry, the low-light images directly inherit the bounding-box and instance-mask annotations of their corresponding normal-light images. As illustrated in Fig.~\ref{fig:paired_annotation_visualization}, the normal-light source images and their synthetic low-light counterparts retain identical defect locations, instance masks, and axis-aligned bounding boxes. The resulting image pairs were used for training and full-reference evaluation of the image enhancement model, whereas the synthesized low-light test images and their shared annotations were used for controlled evaluation of object detection and instance segmentation under low-light conditions.

\subsection{Implementation details}
\label{subsec:Implementation details}

\begin{table*}[pos=h]
\centering
\caption{Architecture of the proposed DaL-MoE and unified training configurations of the compared low-light image enhancement methods and downstream detector architectures.}
\label{tab:training_configs}

\begingroup
\fontsize{5.95}{6.75}\selectfont
\setlength{\tabcolsep}{2.35pt}
\renewcommand{\arraystretch}{1.08}
\setlength{\aboverulesep}{0.28ex}
\setlength{\belowrulesep}{0.28ex}
\setlength{\cmidrulesep}{0.10ex}

\begin{threeparttable}


\begin{minipage}{\linewidth}
\raggedright
\textbf{Panel A. Architecture configuration of the proposed DaL-MoE}
\end{minipage}

\vspace{1.4pt}

\begingroup
\renewcommand{\arraystretch}{1.00}
\setlength{\tabcolsep}{2.0pt}

\begin{tabularx}{\linewidth}{
@{}
>{\RaggedRight\arraybackslash}p{0.48\linewidth}
>{\centering\arraybackslash}p{0.16\linewidth}
>{\centering\arraybackslash}X
@{}}

\toprule

\textbf{Block}
&
\textbf{Input size}
&
\textbf{Output size}
\\

\midrule

Light-up image $I_0$
&
$H\times W\times 3$
&
$H\times W\times 3$
\\

Feature embedding ($3\times3$ Conv)
&
$H\times W\times 3$
&
$H\times W\times 40$
\\

MoE Block (Encoder-1)
&
$H\times W\times 40$
&
$H\times W\times 40$
\\

Downsampling
&
$H\times W\times 40$
&
$H/2\times W/2\times 80$
\\

MoE Block (Encoder-2)
&
$H/2\times W/2\times 80$
&
$H/2\times W/2\times 80$
\\

Downsampling
&
$H/2\times W/2\times 80$
&
$H/4\times W/4\times 160$
\\

MoE Block (Bottleneck)
&
$H/4\times W/4\times 160$
&
$H/4\times W/4\times 160$
\\

Upsampling
&
$H/4\times W/4\times 160$
&
$H/2\times W/2\times 80$
\\

MoE Block (Decoder-2)
&
$H/2\times W/2\times 80$
&
$H/2\times W/2\times 80$
\\

Upsampling
&
$H/2\times W/2\times 80$
&
$H\times W\times 40$
\\

MoE Block (Decoder-1)
&
$H\times W\times 40$
&
$H\times W\times 40$
\\

Output projection and residual reconstruction
&
$H\times W\times 40$
&
$H\times W\times 3$
\\

\bottomrule
\end{tabularx}

\endgroup

\vspace{3.0pt}


\begin{minipage}{\linewidth}
\raggedright
\textbf{Panel B. Principal training settings of the compared low-light image enhancement methods}
\end{minipage}

\vspace{1.4pt}

\begin{tabular*}{\linewidth}{
@{\extracolsep{\fill}}
>{\RaggedRight\arraybackslash}l
>{\centering\arraybackslash}c
>{\centering\arraybackslash}c
>{\centering\arraybackslash}c
>{\centering\arraybackslash}c
>{\centering\arraybackslash}c
>{\centering\arraybackslash}c
>{\centering\arraybackslash}c
@{}}

\toprule

\multirow{2}{*}{\textbf{Method}}
&
\multicolumn{3}{c}{\textbf{Training schedule}}
&
\multicolumn{4}{c}{\textbf{Optimization}}
\\

\cmidrule(lr){2-4}
\cmidrule(lr){5-8}

&
\makecell[c]{\textbf{Patch}\\[-0.5pt]\textbf{size}}
&
\textbf{Batch}
&
\textbf{Iterations}
&
\textbf{Optimizer}
&
$\boldsymbol{LR_0}$
&
$\boldsymbol{LR_{\min}}$
&
\textbf{Schedule}
\\

\midrule

MIRNet~\cite{zamir2020mirnet}
&
$256\times256$
&
8
&
$3\times10^{5}$
&
Adam
&
$2\times10^{-4}$
&
$1\times10^{-6}$
&
Cosine
\\

Restormer~\cite{Zamir2021Restormer}
&
$256\times256$
&
8
&
$3\times10^{5}$
&
Adam
&
$2\times10^{-4}$
&
$1\times10^{-6}$
&
Cosine
\\

LLFormer~\cite{wang2023ultra}
&
$256\times256$
&
8
&
$3\times10^{5}$
&
Adam
&
$2\times10^{-4}$
&
$1\times10^{-6}$
&
Cosine
\\

FourLLIE~\cite{wang2023fourllie}
&
$256\times256$
&
8
&
$3\times10^{5}$
&
Adam
&
$2\times10^{-4}$
&
$1\times10^{-6}$
&
Cosine
\\

Retinexformer~\cite{retinexformer}
&
$256\times256$
&
8
&
$3\times10^{5}$
&
Adam
&
$2\times10^{-4}$
&
$1\times10^{-6}$
&
Cosine
\\

UHDformer~\cite{wang2024uhdformer}
&
$256\times256$
&
8
&
$3\times10^{5}$
&
Adam
&
$2\times10^{-4}$
&
$1\times10^{-6}$
&
Cosine
\\

HVI-CIDNet~\cite{yan2025hvi}
&
$256\times256$
&
8
&
$3\times10^{5}$
&
Adam
&
$2\times10^{-4}$
&
$1\times10^{-6}$
&
Cosine
\\

DarkIR~\cite{Feijoo_2025_CVPR}
&
$256\times256$
&
8
&
$3\times10^{5}$
&
Adam
&
$2\times10^{-4}$
&
$1\times10^{-6}$
&
Cosine
\\

\midrule

\rowcolor{oursbg}
\textbf{DaL-MoE}
&
$256\times256$
&
8
&
$3\times10^{5}$
&
Adam
&
$2\times10^{-4}$
&
$1\times10^{-6}$
&
Cosine
\\

\bottomrule
\end{tabular*}

\vspace{3.0pt}


\begin{minipage}{\linewidth}
\raggedright
\textbf{Panel C. Principal training settings of the downstream detector architectures}
\end{minipage}

\vspace{1.4pt}

\begin{tabular*}{\linewidth}{
@{\extracolsep{\fill}}
>{\RaggedRight\arraybackslash}l
>{\centering\arraybackslash}c
>{\centering\arraybackslash}c
>{\centering\arraybackslash}c
>{\centering\arraybackslash}c
>{\centering\arraybackslash}c
>{\centering\arraybackslash}c
>{\centering\arraybackslash}c
>{\centering\arraybackslash}c
@{}}

\toprule

\multirow{2}{*}{\textbf{Detector}}
&
\multicolumn{3}{c}{\textbf{Training schedule}}
&
\multicolumn{5}{c}{\textbf{Optimization}}
\\

\cmidrule(lr){2-4}
\cmidrule(lr){5-9}

&
\makecell[c]{\textbf{Input}\\[-0.5pt]\textbf{size}}
&
\makecell[c]{\textbf{Effective}\\[-0.5pt]\textbf{batch}}
&
\textbf{Epochs}
&
\textbf{Optimizer}
&
$\boldsymbol{LR_0}$
&
\textbf{Momentum}
&
\textbf{WD}
&
\textbf{Schedule}
\\

\midrule

YOLOv8m~\cite{10533619}
&
$640\times 640$
&
16
&
200
&
SGD
&
0.01
&
0.937
&
$5\times10^{-4}$
&
Cosine
\\

YOLOv11m~\cite{khanam2024yolov11}
&
$640\times640$
&
16
&
200
&
SGD
&
0.01
&
0.937
&
$5\times10^{-4}$
&
Cosine
\\

YOLOv8n~\cite{10533619}
&
$640\times640$
&
16
&
200
&
SGD
&
0.01
&
0.937
&
$5\times10^{-4}$
&
Cosine
\\

YOLOv11n~\cite{khanam2024yolov11}
&
$640\times640$
&
16
&
200
&
SGD
&
0.01
&
0.937
&
$5\times10^{-4}$
&
Cosine
\\

SCNet~\cite{vu2019cascade}
&
$640\times640$
&
16
&
200
&
SGD
&
0.01
&
0.900
&
$1\times10^{-4}$
&
Cosine
\\

\bottomrule
\end{tabular*}

\vspace{1.2pt}

\begin{tablenotes}[flushleft]
\fontsize{5.25}{6.05}\selectfont
\item[]
$H$ and $W$ denote spatial dimensions; $LR_0$, $LR_{\min}$, and WD denote the initial learning rate, terminal learning rate, and weight decay, respectively. All enhancement methods use the same paired training set and augmentation protocol. All detectors are trained on the same normal-light split and evaluated using fixed checkpoints without enhancement-specific fine-tuning. The YOLO entries denote instance-segmentation variants providing both box and mask predictions.
\end{tablenotes}

\end{threeparttable}
\endgroup

\vspace{-4pt}
\end{table*}

We implement DaL-MoE using PyTorch. For a controlled comparison, all image enhancement methods are trained under a unified protocol. Specifically, each model is optimized using Adam for $3\times10^{5}$ iterations. The initial learning rate is set to $2\times10^{-4}$ and gradually decayed to $1\times10^{-6}$ following a cosine annealing schedule. Prior to training, offline data augmentation is performed on the spatially aligned low-light and normal-light image pairs. Random rotations and horizontal and vertical flips are synchronously applied to each input--reference pair to preserve their spatial correspondence, expanding the original 2,413 training image pairs to a total of 12,065 paired training samples. During training, the augmented image pairs are randomly cropped into patches of size $256\times256$, with a batch size of 8. The training objective is to minimize the mean absolute error between the enhanced image and its corresponding normal-light reference image. The downstream detectors are trained and evaluated with an input resolution of 640 × 640. The architecture configuration of DaL-MoE and the unified training settings of the compared image enhancement methods and downstream detector architectures are summarized in Table~\ref{tab:training_configs}.

\subsection{Evaluation Protocol and Metrics}
\label{subsec:evaluation_protocol}

To assess restoration fidelity, downstream bridge damage recognition utility, and transfer to real inspection scenes, three complementary evaluation protocols were adopted. All quantitative evaluations on the paired synthetic benchmark were conducted on the held-out test set containing 482 images. The training--test partition was established before low-light synthesis, ensuring that each normal-light source image and its low-light counterpart remained in the same subset. All compared image enhancement methods were evaluated using the same training and test data.

\paragraph{Full-reference image enhancement evaluation.}
On the paired synthetic test set, peak signal-to-noise ratio (PSNR) and structural similarity index measure (SSIM) were used to compare each enhanced image with its spatially aligned normal-light reference. Following Retinexformer~\cite{retinexformer}, both metrics were calculated on the corresponding RGB image pairs. Let $\hat{I}$ and $I^n$ denote an enhanced image and its normal-light reference, respectively, each with a spatial resolution of $H\times W$ and three color channels. Their mean squared error is defined as
\begin{equation}
    \mathrm{MSE}(\hat{I},I^n)
    =
    \frac{1}{3HW}
    \sum_{p=1}^{HW}
    \sum_{c=1}^{3}
    \left(
        \hat{I}_{p,c}-I^n_{p,c}
    \right)^2.
    \label{eq:evaluation_mse}
\end{equation}
PSNR is subsequently calculated as
\begin{equation}
    \mathrm{PSNR}(\hat{I},I^n)
    =
    10\log_{10}
    \left(
        \frac{L^2}
        {\mathrm{MSE}(\hat{I},I^n)}
    \right),
    \label{eq:evaluation_psnr}
\end{equation}
where $L$ denotes the maximum possible pixel value; $L=1$ for images normalized to $[0,1]$. PSNR is reported in decibels (dB), with a higher value indicating lower pixel-wise reconstruction error.

For corresponding local windows $x$ and $y$ from the enhanced and reference images, respectively, SSIM is defined as
\begin{equation}
    \mathrm{SSIM}(x,y)
    =
    \frac{
        (2\mu_x\mu_y+C_1)
        (2\sigma_{xy}+C_2)
    }{
        (\mu_x^2+\mu_y^2+C_1)
        (\sigma_x^2+\sigma_y^2+C_2)
    },
    \label{eq:evaluation_ssim}
\end{equation}
where $\mu_x$ and $\mu_y$ are the local means,
$\sigma_x^2$ and $\sigma_y^2$ are the local variances, and
$\sigma_{xy}$ is the local covariance. The stabilizing constants are
$C_1=(0.01L)^2$ and $C_2=(0.03L)^2$. The image-level SSIM is obtained by averaging the local-window scores. A higher SSIM indicates greater consistency in luminance, contrast, and structural information. The reported PSNR and SSIM values are arithmetic means over all test image pairs.

\paragraph{Downstream recognition protocol.}
To isolate the effect of image enhancement on downstream recognition, each object detection or instance segmentation model was trained once on the normal-light training split and then held fixed for all input conditions. No enhancement-specific fine-tuning of the downstream models was performed. Each model was evaluated under three input settings: \textit{Normal}, denoting the corresponding normal-light test images as a normal-illumination reference; \textit{Base}, denoting direct inference on the synthesized low-light test images without enhancement; and \textit{Enhanced}, denoting inference on the outputs of the evaluated enhancement methods. The \textit{Normal} setting was not included in the ranking of enhancement methods. The same test images, annotations, detector checkpoints, and inference settings were used across all input conditions; therefore, differences between \textit{Base} and \textit{Enhanced} primarily reflect the influence of the restoration front end.

\paragraph{Object detection and instance segmentation metrics.}
For a predicted region $A$ and a ground-truth region $B$, intersection over union (IoU) is defined as
\begin{equation}
    \mathrm{IoU}(A,B)
    =
    \frac{|A\cap B|}
    {|A\cup B|}.
    \label{eq:evaluation_iou}
\end{equation}
For object detection, $A$ and $B$ denote the predicted and ground-truth bounding boxes, whereas for instance segmentation they denote the predicted and ground-truth masks. At a specified IoU threshold, a prediction is counted as a true positive (TP) when it has the correct class and matches an unmatched ground-truth instance with an IoU not lower than the threshold. Duplicate or unmatched predictions are counted as false positives (FP), and unmatched ground-truth instances are counted as false negatives (FN). Precision and recall are defined as
\begin{equation}
    \mathrm{Precision}
    =
    \frac{\mathrm{TP}}
    {\mathrm{TP}+\mathrm{FP}},
    \qquad
    \mathrm{Recall}
    =
    \frac{\mathrm{TP}}
    {\mathrm{TP}+\mathrm{FN}}.
    \label{eq:evaluation_precision_recall}
\end{equation}

Varying the prediction-confidence threshold produces a precision--recall curve, and the area under its interpolated form gives the average precision (AP) for a given class and IoU threshold. Let $C=8$ denote the number of defect categories and let $\mathrm{AP}_{c,\tau}$ denote the AP of class $c$ at IoU threshold $\tau$. The mean AP at an IoU threshold of 0.50 is
\begin{equation}
    \mathrm{mAP}_{50}
    =
    \frac{1}{C}
    \sum_{c=1}^{C}
    \mathrm{AP}_{c,0.50}.
    \label{eq:evaluation_map50}
\end{equation}
The stricter multi-threshold metric is defined as
\begin{equation}
    \mathrm{mAP}_{50{:}95}
    =
    \frac{1}{C|\mathcal{T}|}
    \sum_{\tau\in\mathcal{T}}
    \sum_{c=1}^{C}
    \mathrm{AP}_{c,\tau},
    \qquad
    \mathcal{T}
    =
    \{0.50,0.55,\ldots,0.95\},
    \label{eq:evaluation_map5095}
\end{equation}
where $|\mathcal{T}|=10$. Thus, $\mathrm{mAP}_{50{:}95}$ averages AP over all eight defect categories and ten IoU thresholds ranging from 0.50 to 0.95 in increments of 0.05. We report box $\mathrm{mAP}_{50}$ and box $\mathrm{mAP}_{50{:}95}$ using bounding-box IoU, together with mask $\mathrm{mAP}_{50}$ and mask $\mathrm{mAP}_{50{:}95}$ using mask IoU. Higher values indicate better classification and localization or instance-boundary delineation performance.

\paragraph{Real-world evaluation.}
The 200 real-world nighttime UAV images do not have paired normal-light references; consequently, full-reference PSNR and SSIM cannot be computed. Dataset-averaged intensity and Canny edge density were therefore used as auxiliary descriptors of visibility changes after enhancement. Let $Y_i$ denote the grayscale intensity image of the $i$-th sample. Its mean intensity is
\begin{equation}
    \mu_i
    =
    \frac{1}{H_iW_i}
    \sum_{p=1}^{H_iW_i}
    Y_i(p).
    \label{eq:evaluation_mean_intensity}
\end{equation}
The dataset-level mean intensity is obtained by averaging $\mu_i$ over all real-world images. Let $E_i$ denote the binary Canny edge map generated using fixed parameters, where edge and non-edge pixels take values of 1 and 0, respectively. The edge density is defined as
\begin{equation}
    d_i
    =
    \frac{1}{H_iW_i}
    \sum_{p=1}^{H_iW_i}
    E_i(p).
    \label{eq:evaluation_edge_density}
\end{equation}
The same grayscale conversion and Canny settings were used for the original and enhanced images. The downstream comparison was also performed using the same fixed YOLOv11m checkpoint and inference settings. Mean intensity and edge density are descriptive rather than full-reference fidelity measures: increased brightness does not necessarily indicate more accurate restoration, and increased edge density may include residual noise or enhancement artifacts. These quantities were therefore excluded from the primary method ranking, and the real-world results were used as complementary evidence of synthetic-to-real transfer.

\begin{table}[pos=h]
\centering
\caption{Comparison of representative low-light image enhancement methods.}
\label{tab:enhancement}

\begingroup
\fontsize{7.4}{8.6}\selectfont
\setlength{\tabcolsep}{5.2pt}
\renewcommand{\arraystretch}{1.10}
\setlength{\aboverulesep}{0.36ex}
\setlength{\belowrulesep}{0.36ex}

\begin{threeparttable}
\begin{tabular*}{0.96\linewidth}{
@{\extracolsep{\fill}}
>{\RaggedRight\arraybackslash}l
>{\centering\arraybackslash}c
>{\centering\arraybackslash}r
>{\centering\arraybackslash}r
@{}}
\toprule
\textbf{Method}
&
\textbf{Venue}
&
\makecell[c]{\textbf{PSNR}\\[-1pt]\textbf{(dB)} $\uparrow$}
&
\makecell[c]{\textbf{SSIM}\\[-1pt]$\uparrow$}
\\
\midrule

MIRNet~\cite{zamir2020mirnet}
&
ECCV'20
&
20.2218
&
0.7903
\\

Restormer~\cite{Zamir2021Restormer}
&
CVPR'22
&
20.6620
&
0.7624
\\

LLFormer~\cite{wang2023ultra}
&
AAAI'23
&
20.7207
&
0.7717
\\

FourLLIE~\cite{wang2023fourllie}
&
ACM MM'23
&
21.1467
&
0.7490
\\

Retinexformer~\cite{retinexformer}
&
ICCV'23
&
21.9448
&
0.7565
\\

UHDformer~\cite{wang2024uhdformer}
&
AAAI'24
&
21.3763
&
0.7183
\\

HVI-CIDNet~\cite{yan2025hvi}
&
CVPR'25
&
21.6674
&
0.7458
\\

DarkIR~\cite{Feijoo_2025_CVPR}
&
CVPR'25
&
21.6787
&
0.7598
\\

\arrayrulecolor{black!22}
\midrule
\arrayrulecolor{black}

\rowcolor{oursbg}
\textbf{Ours}
&
--
&
\textbf{23.1174}
&
\textbf{0.8482}
\\

\bottomrule
\end{tabular*}

\vspace{1.2pt}
\begin{tablenotes}[flushleft]
\fontsize{6.7}{7.8}\selectfont
\item[]
Best results are shown in bold. The green row highlights the proposed method.
\end{tablenotes}

\end{threeparttable}
\endgroup

\vspace{-4pt}
\end{table}

\begin{table*}[pos=h]
\centering
\caption{Downstream bridge damage detection and instance segmentation performance across different detector architectures and image enhancement methods.}
\label{tab:detection}

\begingroup
\fontsize{7.2}{8.4}\selectfont
\setlength{\tabcolsep}{2.7pt}
\renewcommand{\arraystretch}{1.07}
\setlength{\aboverulesep}{0.36ex}
\setlength{\belowrulesep}{0.36ex}
\setlength{\cmidrulesep}{0.14ex}

\begin{threeparttable}
\begin{tabular*}{\linewidth}{
@{\extracolsep{\fill}}
>{\RaggedRight\arraybackslash}l
>{\centering\arraybackslash}l
>{\centering\arraybackslash}c
>{\centering\arraybackslash}c
>{\centering\arraybackslash}c
>{\centering\arraybackslash}c
>{\centering\arraybackslash}c
>{\centering\arraybackslash}c
>{\centering\arraybackslash}c
@{}}

\toprule

\multirow{2}{*}{\makecell[l]{\textbf{Detector}}}
&
\multirow{2}{*}{\makecell[c]{\textbf{Task}}}
&
\multirow{2}{*}{\makecell[c]{\textbf{Metric}}}
&
\multicolumn{2}{c}{\textbf{Reference Inputs}}
&
\multicolumn{4}{c}{\textbf{Enhancement Methods}}
\\

\cmidrule(lr){4-5}
\cmidrule(l{0.10em}r{0.10em}){6-9}

&
&
&
\textbf{Base}
&
\textit{Normal}
&
\makecell[c]{\textbf{Retinexformer}\\[-1pt]\cite{retinexformer}}
&
\makecell[c]{\textbf{HVI-CIDNet}\\[-1pt]\cite{yan2025hvi}}
&
\makecell[c]{\textbf{DarkIR}\\[-1pt]\cite{Feijoo_2025_CVPR}}
&
\cellcolor{oursbg}\textbf{Ours}
\\

\midrule


\multirow{2}{*}{YOLOv8m~\cite{10533619}}
&
\makecell[c]{Box\\[-0.5pt]detection}
&
\makecell[c]{
    $\mathrm{mAP}_{50}$\\[-0.5pt]
    $\mathrm{mAP}_{50{:}95}$
}
&
\makecell[c]{
    0.2787\\[-0.5pt]
    0.1837
}
&
\makecell[c]{
    0.6710\\[-0.5pt]
    0.4744
}
&
\makecell[c]{
    0.4336\\[-0.5pt]
    0.2745
}
&
\makecell[c]{
    0.4193\\[-0.5pt]
    0.2656
}
&
\makecell[c]{
    0.3908\\[-0.5pt]
    0.2466
}
&
\cellcolor{oursbg}
\makecell[c]{
    \textbf{0.4806}\\[-0.5pt]
    \textbf{0.3341}
}
\\

\arrayrulecolor{black!22}
\cmidrule(lr){2-9}
\arrayrulecolor{black}

&
\makecell[c]{Mask\\[-0.5pt]segmentation}
&
\makecell[c]{
    $\mathrm{mAP}_{50}$\\[-0.5pt]
    $\mathrm{mAP}_{50{:}95}$
}
&
\makecell[c]{
    0.1985\\[-0.5pt]
    0.0853
}
&
\makecell[c]{
    0.3913\\[-0.5pt]
    0.1644
}
&
\makecell[c]{
    0.2721\\[-0.5pt]
    0.1169
}
&
\makecell[c]{
    0.2704\\[-0.5pt]
    0.1105
}
&
\makecell[c]{
    0.2291\\[-0.5pt]
    0.0953
}
&
\cellcolor{oursbg}
\makecell[c]{
    \textbf{0.2863}\\[-0.5pt]
    \textbf{0.1251}
}
\\

\midrule


\multirow{2}{*}{YOLOv11m~\cite{khanam2024yolov11}}
&
\makecell[c]{Box\\[-0.5pt]detection}
&
\makecell[c]{
    $\mathrm{mAP}_{50}$\\[-0.5pt]
    $\mathrm{mAP}_{50{:}95}$
}
&
\makecell[c]{
    0.3097\\[-0.5pt]
    0.2105
}
&
\makecell[c]{
    0.7723\\[-0.5pt]
    0.5855
}
&
\makecell[c]{
    0.4495\\[-0.5pt]
    0.2899
}
&
\makecell[c]{
    0.4232\\[-0.5pt]
    0.2701
}
&
\makecell[c]{
    0.3140\\[-0.5pt]
    0.1904
}
&
\cellcolor{oursbg}
\makecell[c]{
    \textbf{0.4923}\\[-0.5pt]
    \textbf{0.3578}
}
\\

\arrayrulecolor{black!22}
\cmidrule(lr){2-9}
\arrayrulecolor{black}

&
\makecell[c]{Mask\\[-0.5pt]segmentation}
&
\makecell[c]{
    $\mathrm{mAP}_{50}$\\[-0.5pt]
    $\mathrm{mAP}_{50{:}95}$
}
&
\makecell[c]{
    0.2281\\[-0.5pt]
    0.1051
}
&
\makecell[c]{
    0.5236\\[-0.5pt]
    0.2409
}
&
\makecell[c]{
    0.2985\\[-0.5pt]
    0.1260
}
&
\makecell[c]{
    0.2836\\[-0.5pt]
    0.1209
}
&
\makecell[c]{
    0.2061\\[-0.5pt]
    0.0844
}
&
\cellcolor{oursbg}
\makecell[c]{
    \textbf{0.3529}\\[-0.5pt]
    \textbf{0.1655}
}
\\

\midrule


\multirow{2}{*}{YOLOv8n~\cite{10533619}}
&
\makecell[c]{Box\\[-0.5pt]detection}
&
\makecell[c]{
    $\mathrm{mAP}_{50}$\\[-0.5pt]
    $\mathrm{mAP}_{50{:}95}$
}
&
\makecell[c]{
    0.3267\\[-0.5pt]
    0.2210
}
&
\makecell[c]{
    0.5720\\[-0.5pt]
    0.3759
}
&
\makecell[c]{
    0.3611\\[-0.5pt]
    0.2169
}
&
\makecell[c]{
    0.3606\\[-0.5pt]
    0.2191
}
&
\makecell[c]{
    0.3099\\[-0.5pt]
    0.1815
}
&
\cellcolor{oursbg}
\makecell[c]{
    \textbf{0.4167}\\[-0.5pt]
    \textbf{0.2718}
}
\\

\arrayrulecolor{black!22}
\cmidrule(lr){2-9}
\arrayrulecolor{black}

&
\makecell[c]{Mask\\[-0.5pt]segmentation}
&
\makecell[c]{
    $\mathrm{mAP}_{50}$\\[-0.5pt]
    $\mathrm{mAP}_{50{:}95}$
}
&
\makecell[c]{
    0.2161\\[-0.5pt]
    0.0961
}
&
\makecell[c]{
    0.3116\\[-0.5pt]
    0.1240
}
&
\makecell[c]{
    0.2085\\[-0.5pt]
    0.0844
}
&
\makecell[c]{
    0.1987\\[-0.5pt]
    0.0828
}
&
\makecell[c]{
    0.1958\\[-0.5pt]
    0.0749
}
&
\cellcolor{oursbg}
\makecell[c]{
    \textbf{0.2430}\\[-0.5pt]
    \textbf{0.1031}
}
\\

\midrule


\multirow{2}{*}{YOLOv11n~\cite{khanam2024yolov11}}
&
\makecell[c]{Box\\[-0.5pt]detection}
&
\makecell[c]{
    $\mathrm{mAP}_{50}$\\[-0.5pt]
    $\mathrm{mAP}_{50{:}95}$
}
&
\makecell[c]{
    0.2764\\[-0.5pt]
    0.1763
}
&
\makecell[c]{
    0.5493\\[-0.5pt]
    0.3556
}
&
\makecell[c]{
    0.3807\\[-0.5pt]
    0.2350
}
&
\makecell[c]{
    0.3508\\[-0.5pt]
    0.2192
}
&
\makecell[c]{
    0.2997\\[-0.5pt]
    0.1803
}
&
\cellcolor{oursbg}
\makecell[c]{
    \textbf{0.4063}\\[-0.5pt]
    \textbf{0.2729}
}
\\

\arrayrulecolor{black!22}
\cmidrule(lr){2-9}
\arrayrulecolor{black}

&
\makecell[c]{Mask\\[-0.5pt]segmentation}
&
\makecell[c]{
    $\mathrm{mAP}_{50}$\\[-0.5pt]
    $\mathrm{mAP}_{50{:}95}$
}
&
\makecell[c]{
    0.1788\\[-0.5pt]
    0.0819
}
&
\makecell[c]{
    0.2952\\[-0.5pt]
    0.1133
}
&
\makecell[c]{
    0.2234\\[-0.5pt]
    0.0901
}
&
\makecell[c]{
    0.2107\\[-0.5pt]
    0.0852
}
&
\makecell[c]{
    0.1912\\[-0.5pt]
    0.0804
}
&
\cellcolor{oursbg}
\makecell[c]{
    \textbf{0.2432}\\[-0.5pt]
    \textbf{0.1027}
}
\\

\midrule


\multirow{2}{*}{SCNet~\cite{vu2019cascade}}
&
\makecell[c]{Box\\[-0.5pt]detection}
&
\makecell[c]{
    $\mathrm{mAP}_{50}$\\[-0.5pt]
    $\mathrm{mAP}_{50{:}95}$
}
&
\makecell[c]{
    0.1115\\[-0.5pt]
    0.0474
}
&
\makecell[c]{
    0.3682\\[-0.5pt]
    0.1688
}
&
\makecell[c]{
    0.2141\\[-0.5pt]
    0.0934
}
&
\makecell[c]{
    0.1783\\[-0.5pt]
    0.0788
}
&
\makecell[c]{
    0.1748\\[-0.5pt]
    0.0762
}
&
\cellcolor{oursbg}
\makecell[c]{
    \textbf{0.2242}\\[-0.5pt]
    \textbf{0.1025}
}
\\

\arrayrulecolor{black!22}
\cmidrule(lr){2-9}
\arrayrulecolor{black}

&
\makecell[c]{Mask\\[-0.5pt]segmentation}
&
\makecell[c]{
    $\mathrm{mAP}_{50}$\\[-0.5pt]
    $\mathrm{mAP}_{50{:}95}$
}
&
\makecell[c]{
    0.1067\\[-0.5pt]
    0.0399
}
&
\makecell[c]{
    0.3594\\[-0.5pt]
    0.1403
}
&
\makecell[c]{
    0.2109\\[-0.5pt]
    0.0832
}
&
\makecell[c]{
    0.1777\\[-0.5pt]
    0.0704
}
&
\makecell[c]{
    0.1742\\[-0.5pt]
    0.0685
}
&
\cellcolor{oursbg}
\makecell[c]{
    \textbf{0.2202}\\[-0.5pt]
    \textbf{0.0873}
}
\\

\bottomrule
\end{tabular*}

\vspace{1.2pt}

\begin{tablenotes}[flushleft]
\fontsize{6.7}{7.8}\selectfont
\item[]
\textit{Base} denotes direct inference on low-light images without enhancement,
whereas \textit{Normal} denotes inference on the corresponding normal-light
images and is excluded from ranking. Within each cell, the upper and lower
values correspond to $\mathrm{mAP}_{50}$ and $\mathrm{mAP}_{50{:}95}$,
respectively. Best results among the enhancement methods are shown in bold.
The green cells highlight the proposed method.
\end{tablenotes}

\end{threeparttable}
\endgroup

\vspace{-4pt}
\end{table*}

\subsection{Comparison with State-of-the-Art Methods}

For the image enhancement task, we compare our method with eight state-of-the-art low-light image enhancement approaches, including MIRNet~\cite{zamir2020mirnet}, Restormer~\cite{Zamir2021Restormer}, LLFormer~\cite{wang2023ultra}, FourLLIE~\cite{wang2023fourllie}, Retinexformer~\cite{retinexformer}, UHDformer~\cite{wang2024uhdformer}, HVI-CIDNet~\cite{yan2025hvi}, and DarkIR~\cite{Feijoo_2025_CVPR}. For downstream bridge damage recognition, we further evaluate the enhanced images using five detector architectures, namely YOLOv8m~\cite{10533619}, YOLOv11m~\cite{khanam2024yolov11}, YOLOv8n~\cite{10533619}, YOLOv11n~\cite{khanam2024yolov11}, and SCNet~\cite{vu2019cascade}. Because each synthetic low-light image is generated from a normal-light source while preserving the original scene geometry and defect annotations, the same bounding boxes and instance masks can be used to compare recognition performance under different illumination conditions. For the downstream comparison reported in Table~\ref{tab:detection}, three representative enhancement baselines, namely Retinexformer, HVI-CIDNet, and DarkIR, are compared with the proposed DaL-MoE. Each detector is trained once on the normal-light training split and then kept fixed for inference on the \textit{Base}, \textit{Normal}, and enhanced inputs. Therefore, the differences in downstream performance primarily reflect the contribution of the restoration front end rather than changes in detector architecture or optimization. In addition, two reference settings are reported: \textit{Base} denotes direct inference on low-light images without enhancement, whereas \textit{Normal} denotes inference on the corresponding normal-light images and serves as an upper-bound reference under normal illumination.

\subsubsection{Quantitative Comparison}

Based on the quantitative results reported in Table~\ref{tab:enhancement} for image enhancement and Table~\ref{tab:detection} for downstream bridge damage recognition, four observations can be made.

(1) \textbf{DaL-MoE achieves the best image enhancement performance on the paired bridge benchmark.} As shown in Table~\ref{tab:enhancement}, DaL-MoE obtains the highest performance among all compared low-light image enhancement methods, achieving a PSNR of $23.1174$~dB and an SSIM of $0.8482$. Retinexformer is the strongest baseline in terms of PSNR, with a PSNR of $21.9448$~dB; therefore, DaL-MoE provides an improvement of $1.1726$~dB. In terms of SSIM, MIRNet is the best-performing baseline with a score of $0.7903$, whereas DaL-MoE improves this value by $0.0579$. These results indicate that the proposed method can simultaneously improve pixel-level reconstruction fidelity and structural consistency. This property is particularly important for low-light UAV bridge inspection, where useful evidence may be distributed across weak crack patterns, spalling contours, corrosion traces, seepage regions, and other low-contrast surface defects. In contrast to methods that may obtain favorable performance on one metric while compromising the other, DaL-MoE provides a more balanced recovery of brightness, color, local contrast, and structural information.

(2) \textbf{Low-light enhancement improves downstream bridge damage recognition.} As reported in Table~\ref{tab:detection}, directly applying the detector to low-light images, denoted as \textit{Base}, results in a clear performance degradation. This degradation is caused by the loss of local contrast, amplified dark-region noise, and weakened defect boundaries under insufficient illumination. In contrast, the images enhanced by DaL-MoE consistently improve both bounding-box detection and instance segmentation performance. For example, on YOLOv8m, DaL-MoE increases box $\mathrm{mAP}_{50}$ from $0.2787$ to $0.4806$ and box $\mathrm{mAP}_{50{:}95}$ from $0.1837$ to $0.3341$. On YOLOv11m, box $\mathrm{mAP}_{50}$ increases from $0.3097$ to $0.4923$, while box $\mathrm{mAP}_{50{:}95}$ increases from $0.2105$ to $0.3578$. These improvements show that the proposed restoration front end can recover visual cues that are useful for defect localization and recognition, rather than merely increasing the apparent brightness of the input images.

(3) \textbf{DaL-MoE provides consistent advantages in both detection and instance segmentation.} Among the enhancement methods reported in Table~\ref{tab:detection}, DaL-MoE achieves the highest values in every reported detector--task combination for both bounding-box detection and mask segmentation. For example, on YOLOv11m, DaL-MoE obtains a box $\mathrm{mAP}_{50}$ of $0.4923$ and a box $\mathrm{mAP}_{50{:}95}$ of $0.3578$, outperforming Retinexformer by $0.0428$ and $0.0679$, respectively. For instance segmentation, DaL-MoE increases mask $\mathrm{mAP}_{50}$ from $0.2985$ to $0.3529$ and mask $\mathrm{mAP}_{50{:}95}$ from $0.1260$ to $0.1655$ compared with Retinexformer. These results suggest that DaL-MoE not only improves global visibility but also preserves local defect structures, object boundaries, and fine-grained surface details that are essential for accurate instance-level delineation. The consistent improvement in both box- and mask-based metrics further supports the task-oriented design of the proposed restoration model.

(4) \textbf{The proposed method provides stable gains across different detector architectures.} The improvements brought by DaL-MoE are not restricted to a particular detector family or model scale. Compared with the \textit{Base} setting, DaL-MoE improves box $\mathrm{mAP}_{50}$ by $0.2019$, $0.1826$, $0.09$, $0.1299$, and $0.1127$ for YOLOv8m, YOLOv11m, YOLOv8n, YOLOv11n, and SCNet, respectively. The same overall trend is also observed for box $\mathrm{mAP}_{50{:}95}$, mask $\mathrm{mAP}_{50}$, and mask $\mathrm{mAP}_{50{:}95}$. These results indicate that the benefit of the proposed enhancement is not tied to a specific detector architecture and supports the use of DaL-MoE as a detector-agnostic restoration front end. Although the \textit{Normal} setting remains an upper-bound reference under the evaluated conditions, DaL-MoE consistently narrows the performance gap between low-light and normal-light inputs. This demonstrates the practical value of task-oriented low-light restoration for UAV-based bridge damage inspection.

\subsubsection{Visual Comparison}

To further evaluate the effectiveness of the proposed method, we provide visual comparisons from two complementary perspectives: enhanced bridge image quality and downstream damage recognition performance.

\begin{figure}[pos=htbp]
    \centering
    \includegraphics[width=\linewidth]{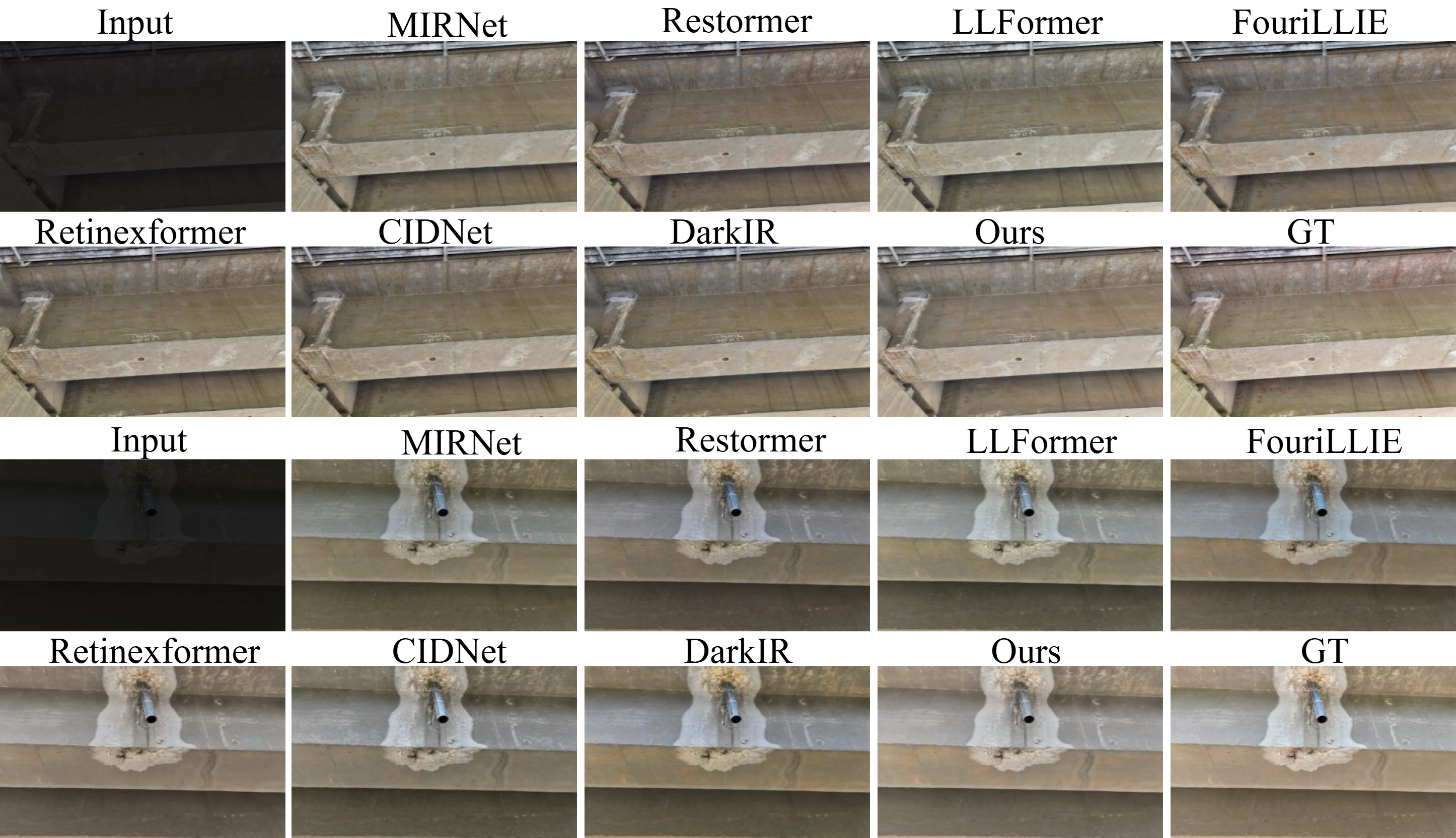}
    \caption{Visual comparison of low-light bridge image enhancement results. \textit{Input} denotes the original low-light image, \textit{GT} denotes the corresponding normal-light reference, and the remaining columns show the enhanced results produced by different methods.}
    \label{fig:visual_enhance}
\end{figure}

\textbf{Visual comparison of enhanced images.} Fig.~\ref{fig:visual_enhance} presents representative visual comparisons among the input low-light UAV images, the corresponding normal-light references, and the enhanced results produced by different methods. Here, \textit{Input} denotes the original low-light image, whereas \textit{GT} denotes the corresponding normal-light reference image. Across multiple challenging bridge scenes, DaL-MoE restores the overall illumination more faithfully toward the normal-light reference, making bridge surfaces and damage-related regions more visible. More importantly, the proposed method maintains stable color rendition and preserves local contrast without introducing obvious chromatic distortion. In contrast, although MIRNet and Restormer improve the overall visibility to some extent, their results exhibit noticeable color deviations in several scenes, leading to unnatural tones and reduced visual consistency with the reference images. These deviations may affect the appearance of concrete textures and defect regions, particularly when the defects are characterized by subtle color or intensity differences. Compared with these methods, DaL-MoE achieves a better balance between brightness recovery, color fidelity, and structural preservation. This indicates that the proposed approach is better suited to enhancing low-light bridge imagery while retaining visual cues that are relevant to subsequent damage recognition.

\begin{figure}[pos=htbp]
    \centering
    \includegraphics[width=\linewidth]{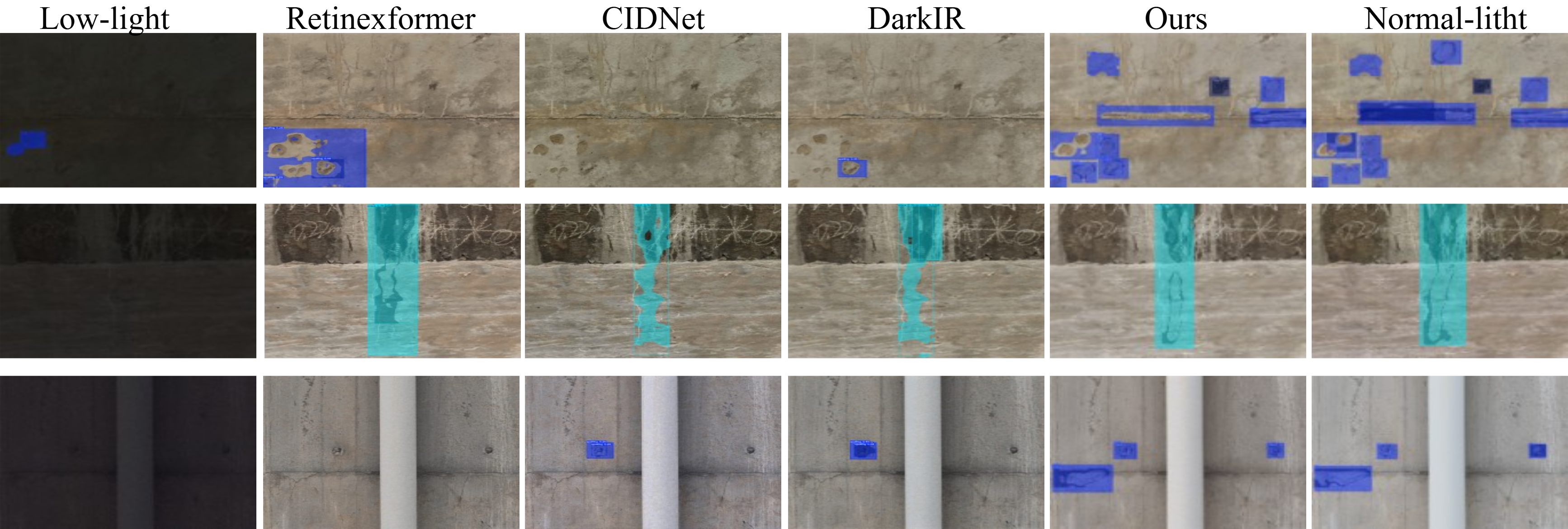}
    \caption{Visual comparison of YOLOv11m bridge damage detection results under different input conditions. The same detector is applied to the low-light input, the enhanced images, and the corresponding normal-light reference.}
    \label{fig:visual_detection}
\end{figure}

\textbf{Visual comparison of downstream detection results.} Fig.~\ref{fig:visual_detection} presents the YOLOv11m detection results under different input settings. Two observations can be made. (1) \textbf{Enhancement before detection substantially improves bridge damage recognition under low-light conditions.} When the original low-light images are directly fed into YOLOv11m, the detector produces noticeable missed detections and incomplete localization results. This is mainly because insufficient illumination weakens defect visibility, reduces local contrast, blurs damage boundaries, and makes small or low-contrast defects difficult to distinguish from the surrounding bridge surface. Such effects are particularly harmful for fine cracks, shallow spalling, corrosion traces, and other defects whose visual evidence is already weak. After low-light enhancement, the visual quality of the input images is improved and more damage-relevant cues become visible. As a result, YOLOv11m detects more defect regions and produces more complete prediction results. This qualitative observation is consistent with the quantitative improvements reported in Table~\ref{tab:detection}, demonstrating that image enhancement can serve as an effective preprocessing stage for low-light UAV bridge damage recognition.

(2) \textbf{DaL-MoE provides more detection-friendly enhanced images than the other compared enhancement methods.} Although some existing methods can produce visually acceptable results in particular scenes, their enhanced outputs are not always equally beneficial for downstream bridge damage recognition. For example, Retinexformer may recover reasonable brightness and visual appearance in some cases, but its enhanced images can still lead to missed detections when processed by YOLOv11m. This observation indicates that perceptual brightness improvement alone is insufficient for bridge damage inspection. The detector also depends on whether damage-relevant cues, such as crack boundaries, spalling contours, corrosion patterns, seepage regions, and local texture variations, are preserved with sufficient fidelity. In contrast, DaL-MoE not only improves image brightness but also preserves damage-relevant structural details while reducing color distortion and boundary degradation. Consequently, the enhanced images generated by DaL-MoE enable YOLOv11m to identify more complete defect regions and produce results closer to those obtained under normal illumination. These results further demonstrate that the proposed method is suitable for task-oriented low-light restoration, where the objective is not merely to improve perceptual quality but to recover visual information that supports accurate downstream bridge damage detection.

\subsection{Ablation Study}

To verify the effectiveness of the proposed components for task-oriented low-light restoration of bridge inspection images, we conduct ablation experiments on the paired synthetic low-light bridge benchmark. The analysis is performed from two perspectives: the overall contribution of the Degradation-Aware Guidance Estimation (DAGE) module and the Adaptive Mixture-of-Experts Restoration (AMER) module, and the internal design of the AMER module. The results are reported in Table~\ref{tab:ablation}. Since the benchmark preserves the spatial correspondence between normal-light and low-light images as well as the original bridge damage annotations, the ablation results reflect the ability of each component to recover illumination, suppress degradation, and preserve damage-relevant structural information.

\begin{table}[pos=h]
\centering
\caption{Ablation study of the proposed DaL-MoE method.}
\label{tab:ablation}

\begingroup
\fontsize{7.2}{8.4}\selectfont
\setlength{\tabcolsep}{3.0pt}
\renewcommand{\arraystretch}{1.12}
\setlength{\aboverulesep}{0.34ex}
\setlength{\belowrulesep}{0.34ex}
\setlength{\cmidrulesep}{0.16ex}

\begin{threeparttable}
\begin{tabular*}{\linewidth}{
@{\extracolsep{\fill}}
>{\RaggedRight\arraybackslash}l
*{6}{>{\centering\arraybackslash}c}
>{\centering\arraybackslash}r
>{\centering\arraybackslash}r
@{}}
\toprule

\multirow{2}{*}{\textbf{Variant}}
&
\multicolumn{2}{c}{\textbf{Main Modules}}
&
\multicolumn{4}{c}{\textbf{AMER Components}}
&
\multicolumn{2}{c}{\textbf{Performance}}
\\

\cmidrule(lr){2-3}
\cmidrule(lr){4-7}
\cmidrule(lr){8-9}

&
\textbf{DAGE}
&
\textbf{AMER}
&
$\boldsymbol{E_1}$
&
$\boldsymbol{E_2}$
&
$\boldsymbol{E_3}$
&
\textbf{Gate}
&
\makecell[c]{\textbf{PSNR}\\[-1pt]\textbf{(dB)} $\uparrow$}
&
\makecell[c]{\textbf{SSIM}\\[-1pt]$\uparrow$}
\\

\midrule

\rowcolor{summarygray}
\multicolumn{9}{@{}l@{}}{%
    \strut\textbf{A. Main architecture modules}%
}
\\[-0.15ex]

w/o AMER
&
\abyes
&
\abno
&
\abdash
&
\abdash
&
\abdash
&
\abdash
&
21.65
&
0.7157
\\

w/o DAGE
&
\abno
&
\abyes
&
\abyes
&
\abyes
&
\abyes
&
\abyes
&
22.73
&
0.7716
\\

\addlinespace[0.7ex]

\rowcolor{summarygray}
\multicolumn{9}{@{}l@{}}{%
    \strut\textbf{B. AMER components}%
}
\\[-0.15ex]

w/o Expert$_1$
&
\abyes
&
\abyes
&
\abno
&
\abyes
&
\abyes
&
\abyes
&
22.03
&
0.7682
\\

w/o Expert$_2$
&
\abyes
&
\abyes
&
\abyes
&
\abno
&
\abyes
&
\abyes
&
22.89
&
0.7449
\\

w/o Expert$_3$
&
\abyes
&
\abyes
&
\abyes
&
\abyes
&
\abno
&
\abyes
&
22.72
&
0.7553
\\

w/o Gate
&
\abyes
&
\abyes
&
\abyes
&
\abyes
&
\abyes
&
\abno
&
21.14
&
0.7397
\\

\addlinespace[0.45ex]

\cellcolor{oursbg}\textbf{Ours}
&
\abyes
&
\abyes
&
\abyes
&
\abyes
&
\abyes
&
\abyes
&
\cellcolor{oursbg}\textbf{23.12}
&
\cellcolor{oursbg}\textbf{0.8482}
\\

\bottomrule
\end{tabular*}

\vspace{1.3pt}
\begin{tablenotes}[flushleft]
\fontsize{6.6}{7.8}\selectfont
\item[]
``w/o'' denotes ``without.'' 
\abyes, \abno, and \abdash indicate enabled, removed, and not applicable, respectively.
DAGE and AMER denote the Degradation-Aware Guidance Estimation module and the Adaptive Mixture-of-Experts Restoration module, respectively.
$E_1$--$E_3$ denote the Noise-Aware Expert, Color Adjustment Expert, and Detail Enhancement Expert, respectively.
The best results are shown in bold, and the green cells highlight the full model and its performance.
\end{tablenotes}

\end{threeparttable}
\endgroup

\vspace{-4pt}
\end{table}

\textbf{Ablation of DAGE and AMER.}
We first evaluate the contribution of the two main modules, namely DAGE and AMER. When AMER is removed, the PSNR and SSIM decrease from $23.12$ and $0.8482$ to $21.65$ and $0.7157$, respectively. This degradation demonstrates that adaptive expert collaboration is important for restoring bridge images affected by coupled low-light degradations, including insufficient illumination, dark-region noise amplification, contrast loss, and weakened damage boundaries. \textit{Without the AMER} module, the network has a reduced capacity to coordinate noise suppression, illumination adjustment, and structural-detail recovery, which are jointly required to recover visual cues from cracks, spalling contours, corrosion traces, seepage regions, and other low-contrast bridge defects. \textit{Removing DAGE} module also results in a clear performance degradation, with the PSNR decreasing to $22.73$ and the SSIM decreasing to $0.7716$. DAGE estimates spatially varying illumination and noise-related guidance from the low-light input and provides this information to the subsequent restoration stages. When DAGE is removed, the expert branches receive less explicit information about regions affected by severe underexposure and noise amplification. As a result, the restoration network becomes less effective in selectively suppressing low-light degradation while preserving damage-relevant structures. These results demonstrate that DAGE and AMER play complementary roles: DAGE provides degradation-aware guidance, whereas AMER uses this guidance to adaptively integrate different restoration behaviors.

\textbf{Ablation of the AMER components.}
We further investigate the internal design of AMER by individually removing the three expert branches and the adaptive gating mechanism while retaining the other components. The three experts correspond to the Noise-Aware Expert, the Color Adjustment Expert, and the Detail Enhancement Expert, which are designed to address complementary degradation characteristics in low-light bridge imagery. \textit{Removing Expert$_1$}, namely the Noise-Aware Expert, reduces the PSNR and SSIM to $22.03$ and $0.7682$, respectively. This result indicates that degradation-aware noise suppression is important for recovering information from dark regions, where sensor noise and illumination loss can obscure weak damage patterns. In particular, effective noise suppression helps prevent noise responses from interfering with the representation of fine cracks, corrosion traces, and other low-contrast defects. \textit{Removing Expert$_2$}, namely the Color Adjustment Expert, produces a PSNR of $22.89$ and an SSIM of $0.7449$. Although its PSNR remains relatively high among the incomplete AMER variants, the lower SSIM indicates a loss of structural and tonal consistency. This result suggests that color and illumination adjustment contributes not only to the visual appearance of the restored image but also to the consistency of bridge-surface structures and damage-related regions with the normal-light reference.
\textit{Removing Expert$_3$}, namely the Detail Enhancement Expert, decreases the PSNR and SSIM to $22.72$ and $0.7553$, respectively. This degradation confirms the importance of the detail-oriented branch for recovering local boundaries and fine structures. Such information is particularly relevant to bridge damage categories with weak or irregular morphology, including thin cracks, spalling boundaries, corrosion contours, exposed reinforcement edges, and local concrete texture variations.

The \textit{adaptive gating mechanism} is also important for the proposed AMER design. When the gate is removed, the model obtains a PSNR of $21.14$ and an SSIM of $0.7397$. The PSNR of this variant is the lowest among all evaluated variants, indicating that simply combining multiple experts without degradation-dependent routing is insufficient. The gating network enables the model to dynamically adjust the contribution of the three experts according to the degradation characteristics of the current input and feature scale. This adaptive routing improves the coordination between noise suppression, color and illumination adjustment, and structural-detail recovery.

Overall, removing any individual expert or the gating mechanism leads to lower performance than the full model. The ablation results therefore confirm that the three experts provide complementary restoration capabilities, while the adaptive gating mechanism is necessary to integrate them effectively. Together with the contribution of DAGE, these results support the task-oriented design of DaL-MoE, which aims not only to improve the perceptual quality of low-light bridge images but also to recover visual information that is useful for subsequent multi-class damage detection and instance segmentation.

\section{Real-World Evaluation}
\label{sec:real_world_evaluation}

\begin{figure}[pos=h]
    \centering
    \includegraphics[width=0.5\linewidth]{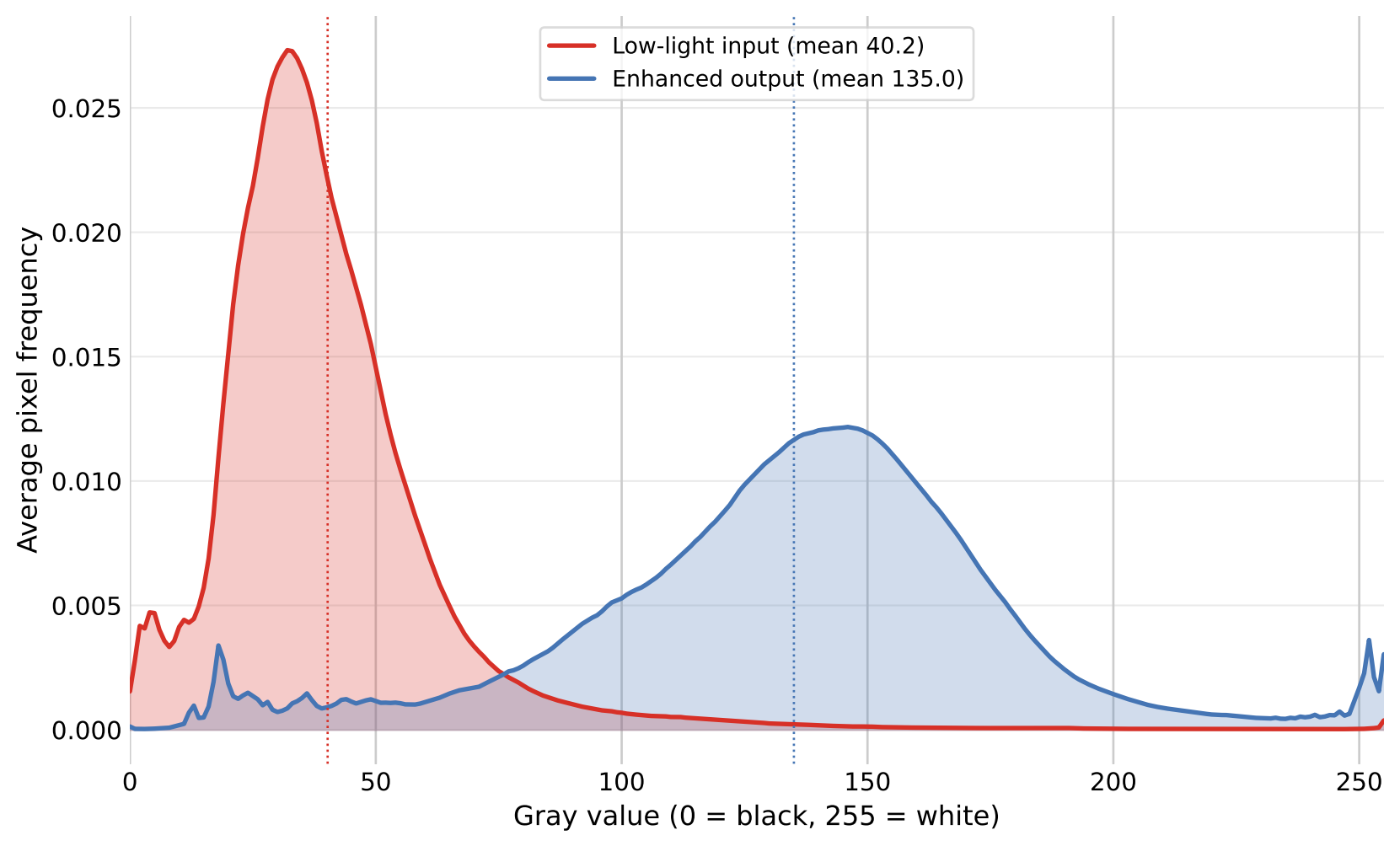}
    \caption{Dataset-averaged grayscale intensity distributions of 200 real-world nighttime bridge images. DaL-MoE increases the mean intensity from 40.2 to 135.0.}
    \label{fig:real_brightness}
\end{figure}

To further examine the practical applicability and synthetic-to-real transfer capability of the proposed DaL-MoE framework, we evaluate it on real-world low-light UAV bridge inspection images. Unlike the paired synthetic benchmark, these images do not have spatially aligned normal-light references. Therefore, full-reference restoration metrics such as PSNR and SSIM cannot be computed for this real-world set. We instead evaluate real-scene transfer from two complementary perspectives: image-level visibility analysis and downstream bridge damage detection. The former characterizes changes in brightness and local edge responses, while the latter includes both qualitative visualization and quantitative box detection using manual annotations. This evaluation examines whether the degradation-aware restoration learned from the ISP-aware synthetic benchmark can transfer to practical bridge inspection scenes affected by severe underexposure, structural occlusion, and illumination variability.

\subsection{Visual Analysis of Real-World Nighttime Bridge Images}

We evaluate 200 real-world nighttime bridge images captured during UAV inspections. The visual analysis focuses on two aspects: whether DaL-MoE alleviates severe underexposure and whether it reveals local structural cues that are difficult to perceive in the original low-light inputs, as illustrated in Figs.~\ref{fig:real_brightness} and~\ref{fig:real_structure}. Since the real-world images do not have paired normal-light references, the following intensity and edge-density measurements serve as auxiliary descriptors of visibility changes rather than full-reference restoration metrics.

\subsubsection{Brightness Enhancement}

We first examine whether DaL-MoE effectively alleviates the severe underexposure commonly observed in real-world nighttime bridge inspection scenes. As shown in Fig.~\ref{fig:real_brightness}, the dataset-averaged grayscale intensity distribution of the original low-light images is concentrated in the dark range, particularly below a gray value of 60. The mean intensity is only 40.2, indicating that a considerable amount of bridge-surface information is poorly exposed and difficult to distinguish from the surrounding background.

After enhancement, the intensity distribution shifts toward a substantially broader and more informative brightness range. The mean intensity increases from 40.2 to 135.0, corresponding to a $3.36\times$ increase. The enhanced distribution does not exhibit a dominant accumulation at the high-intensity end, suggesting that DaL-MoE improves the visibility of underexposed bridge regions without causing obvious global saturation. This brightness recovery is particularly beneficial for surfaces located beneath bridge decks, within locally occluded regions, and around structural components where damage cues may otherwise be submerged in darkness. Together, these observations indicate that the illumination-recovery capability learned from the synthetic low-light benchmark transfers effectively to real nighttime UAV imagery and improves the visibility of underexposed bridge surfaces.

\subsubsection{Structural Cue Recovery}

\begin{figure}[pos=h]
    \centering
    \includegraphics[width=0.5\linewidth]{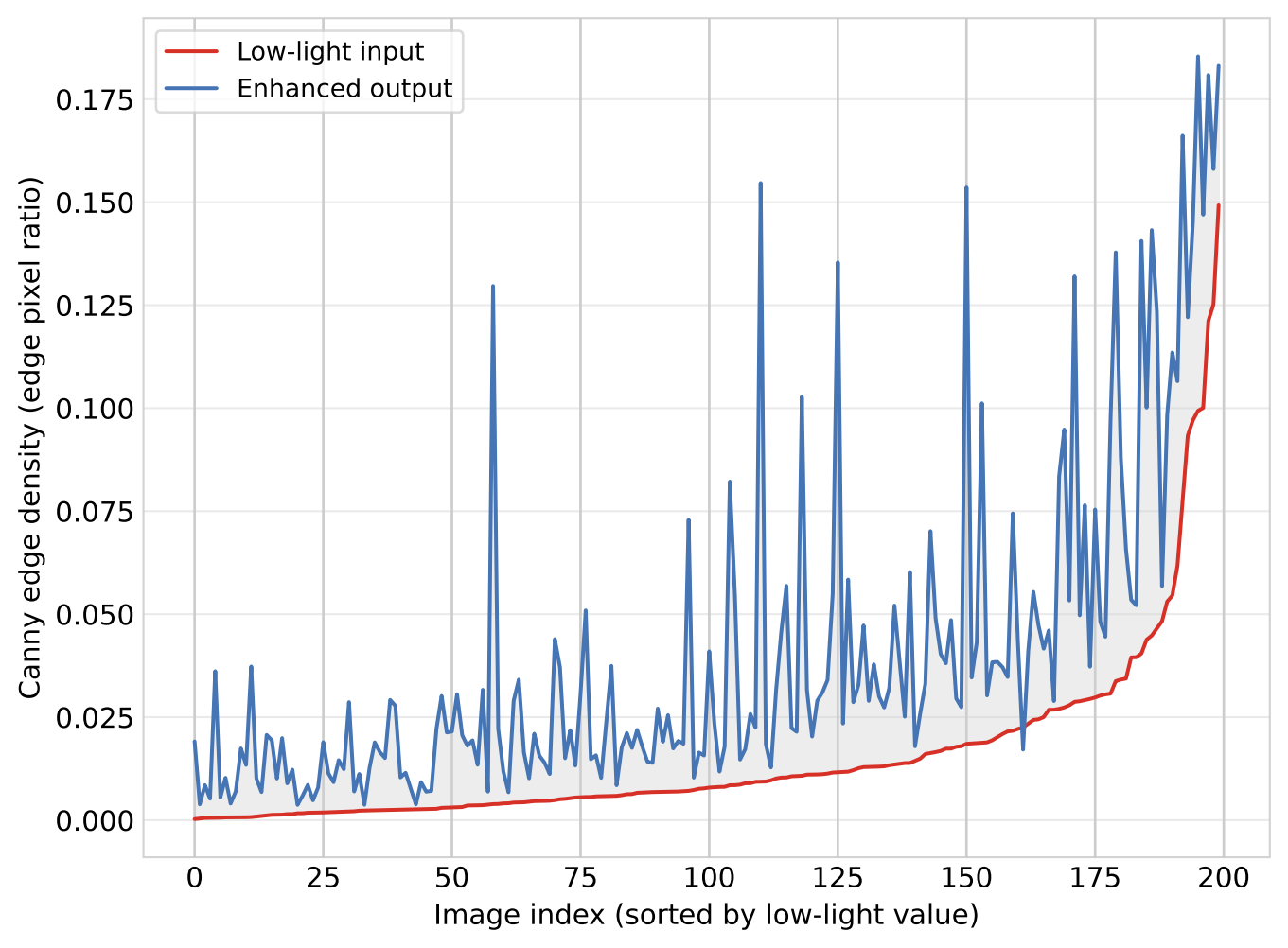}
    \caption{Canny edge densities of 200 real-world nighttime bridge images, sorted by the density of the low-light inputs. The mean density increases from $1.55\%$ to $4.07\%$.}
    \label{fig:real_structure}
\end{figure}

Improved brightness alone does not necessarily indicate that damage-relevant structural information has been recovered. We therefore further analyze local structural visibility using Canny edge density~\cite{Canny1986TPAMI}, defined as the ratio of detected edge pixels to the total number of image pixels. The same grayscale conversion and fixed Canny parameters are applied to each low-light image and its corresponding enhanced output, so that the comparison reflects the change induced by the restoration front end. As shown in Fig.~\ref{fig:real_structure}, the enhanced outputs generally exhibit higher edge density than the corresponding low-light inputs across the 200-image dataset. The dataset-average edge density increases from 1.55\% to 4.07\%. This increase indicates that more local intensity transitions and high-frequency responses become detectable after enhancement, which is consistent with improved visibility of surface boundaries, construction joints, local texture variations, and potential damage regions.

\begin{figure}[pos=h]
    \centering
    \includegraphics[width=0.75\linewidth]{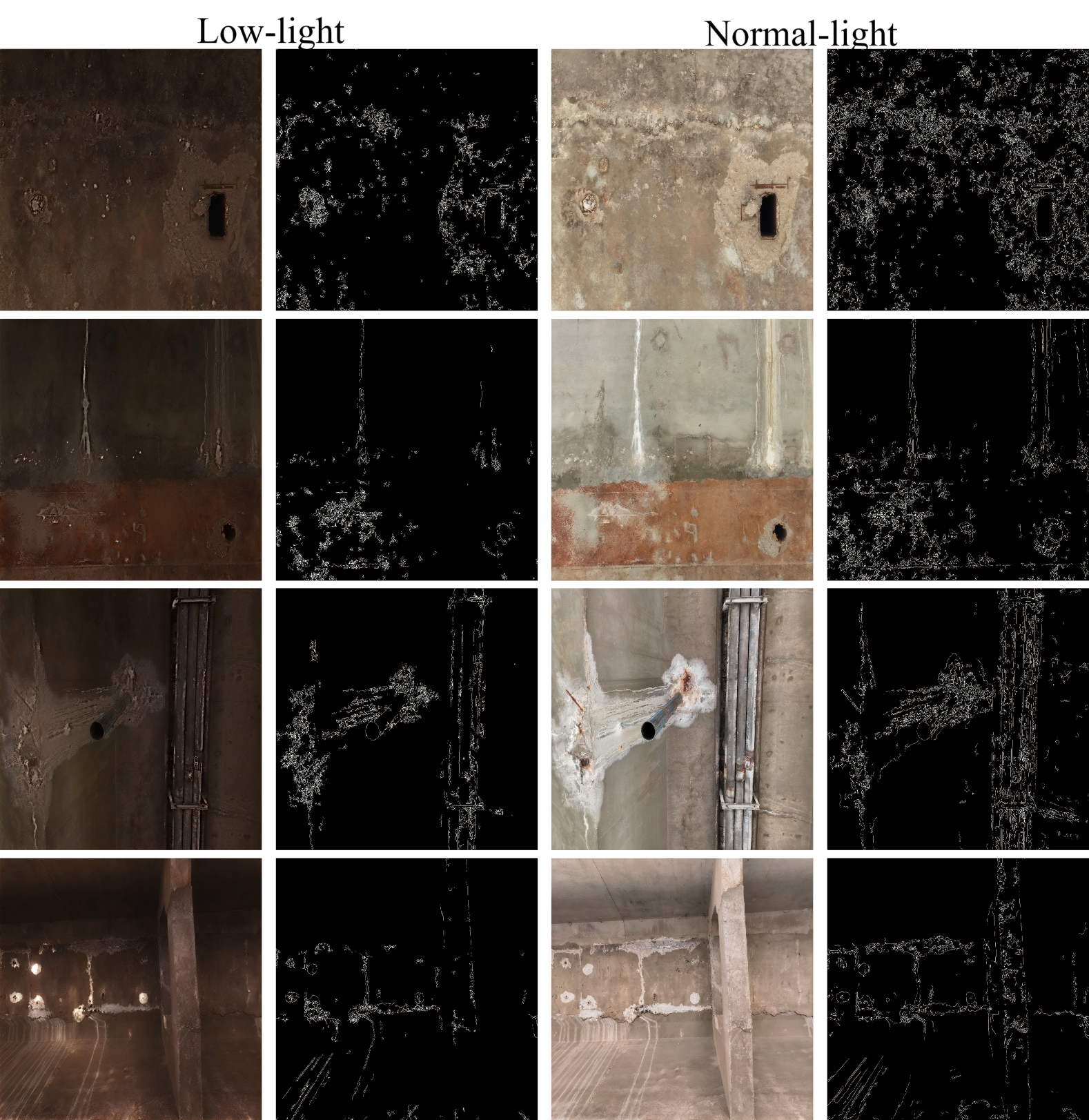}
    \caption{Structural-cue comparison on representative real-world nighttime bridge images. From left to right: low-light input, input Canny edges, DaL-MoE output, and output Canny edges.}
    \label{fig:real_canny}
\end{figure}

However, Canny edge density is an auxiliary visibility descriptor rather than a direct measure of structural fidelity. A higher edge density may reflect the recovery of meaningful boundaries, but it may also include enhanced texture responses or residual noise. For this reason, the edge-density analysis is interpreted together with the qualitative examples and downstream detection results rather than used as an independent measure of restoration accuracy. For a more intuitive comparison, Fig.~\ref{fig:real_canny} presents four representative samples together with their corresponding Canny edge maps. In the original low-light inputs, many surface boundaries and local patterns are weak or fragmented because of insufficient illumination. After enhancement, denser and more continuous edge responses can be observed around structural boundaries, construction joints, surface textures, and damage-related regions such as cracks and spalling. The enhanced edge maps may also contain additional fine responses caused by noise or texture amplification; nevertheless, the overall increase in edge continuity and local responses provides useful qualitative evidence that structural cues have become more visible.

\subsection{Downstream Bridge Damage Detection}

\begin{figure*}[pos=t]
    \centering
    \includegraphics[width=1.0\linewidth]{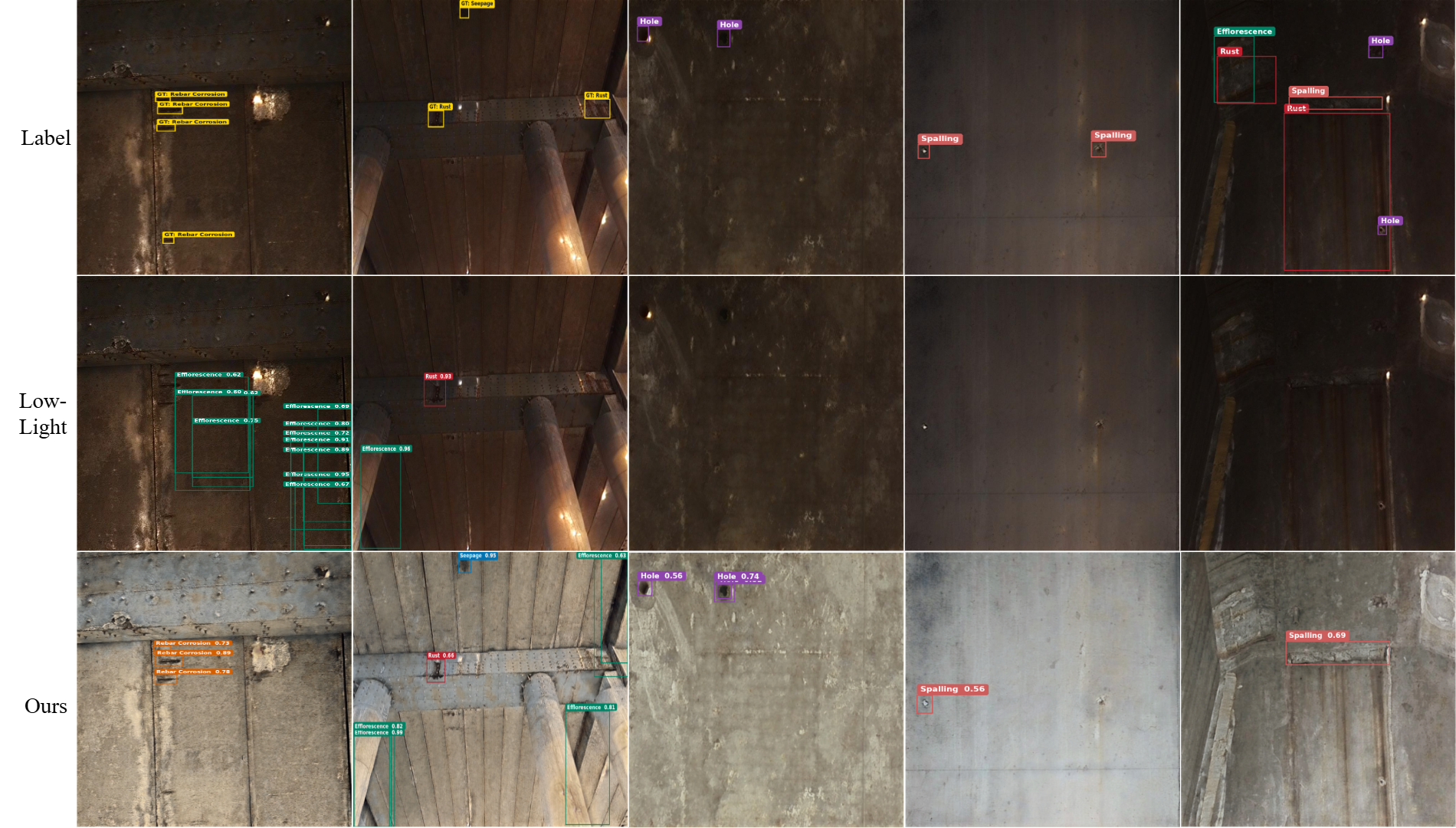}
    \caption{Qualitative YOLOv11m detection results on real-world nighttime bridge images. \textit{Label}, \textit{Low-light}, and \textit{Ours} denote manual annotations, predictions on the original low-light images, and predictions after DaL-MoE enhancement, respectively. The detector and inference settings are fixed.}
    \label{fig:realWorld}
\end{figure*}

The preceding analysis shows that DaL-MoE improves the brightness distribution and exposes richer local edge responses in real-world nighttime bridge images. We next examine whether this recovered visual information can be utilized by a downstream bridge damage detector, thereby providing a real-scene evaluation of the task-oriented restoration objective introduced in this paper. Although paired normal-light references are unavailable, all 200 original low-light images were manually annotated with bounding boxes for the same eight damage categories considered in the paired synthetic benchmark. Because DaL-MoE changes the photometric appearance without altering the spatial geometry of the input images, the same annotations can be used to evaluate detection on both the original low-light images and their enhanced counterparts. This setting enables both qualitative and quantitative evaluation of real-world box detection performance.

As shown in Fig.~\ref{fig:realWorld}, \textit{Label} denotes the bounding-box annotations manually delineated on the original low-light images. \textit{Low-light} denotes direct inference on the original low-light images, while \textit{Ours} denotes inference after enhancement with DaL-MoE. Consistent with the controlled downstream evaluation protocol described in Section~\ref{subsec:evaluation_protocol}, damage detection in both settings is performed using the same YOLOv11m checkpoint trained on the normal-light training split of the collected bridge damage dataset. The detector architecture, model parameters, preprocessing, and inference procedure are kept fixed, and no enhancement-specific fine-tuning is applied. Therefore, differences in the visual predictions and quantitative box metrics primarily reflect the influence of the restoration front end on downstream damage perception.

When the original low-light images are directly fed into YOLOv11m, the detector exhibits missed detections, spurious responses, and inaccurate localization in several challenging examples. Severe underexposure reduces the local contrast between bridge damage and the surrounding concrete surface, while darkness and sensor noise obscure damage boundaries and fine surface textures. These effects make it difficult for the detector to distinguish defects such as cracks, spalling, corrosion-related regions, and other low-contrast surface deteriorations.

In contrast, the detections obtained from the DaL-MoE-enhanced images are generally more consistent with the manual annotations in the illustrated examples. Several previously weak or missed damage regions become detectable after enhancement, and the predicted regions are more closely aligned with visible bridge-surface structures. This improvement is consistent with the preceding image-level analysis: by recovering illumination and exposing local structural cues, DaL-MoE provides a more informative RGB input for the fixed downstream detector. The visual examples also show that some missed detections and spurious predictions remain, reflecting the severity and spatial heterogeneity of real-world low-light degradation as well as the domain gap from the synthetic training distribution. Nevertheless, the comparison indicates that the visual information recovered by DaL-MoE contributes to downstream damage recognition rather than merely improving perceived image brightness.

\begin{table}[pos=t]
\centering
\caption{Class-wise box detection performance and absolute gains on 200 manually annotated real-world nighttime bridge images.}
\label{tab:detection_map}

\begingroup
\fontsize{6.35}{7.45}\selectfont
\setlength{\tabcolsep}{1.7pt}
\renewcommand{\arraystretch}{1.10}
\setlength{\aboverulesep}{0.32ex}
\setlength{\belowrulesep}{0.32ex}
\setlength{\cmidrulesep}{0.12ex}

\begin{threeparttable}
\begin{tabular*}{\linewidth}{
@{\extracolsep{\fill}}
>{\RaggedRight\arraybackslash}l
*{6}{>{\centering\arraybackslash}c}
@{}}

\toprule

\multirow{2}{*}{\textbf{Class}}
&
\multicolumn{2}{c}{\textbf{Low-light}}
&
\multicolumn{2}{c}{\textbf{Enhanced}}
&
\multicolumn{2}{c}{\textbf{Absolute gain}}
\\

\cmidrule(lr){2-3}
\cmidrule(lr){4-5}
\cmidrule(lr){6-7}

&
$\mathrm{AP}_{50}$
&
$\mathrm{AP}_{50{:}95}$
&
$\mathrm{AP}_{50}$
&
$\mathrm{AP}_{50{:}95}$
&
$\Delta\mathrm{AP}_{50}$
&
$\Delta\mathrm{AP}_{50{:}95}$
\\

\midrule

Spalling
&
0.079
&
0.033
&
\textbf{0.240}
&
\textbf{0.109}
&
+0.161
&
+0.076
\\

Crack
&
0.050
&
0.019
&
\textbf{0.125}
&
\textbf{0.057}
&
+0.075
&
+0.038
\\

Hole
&
0.224
&
0.096
&
\textbf{0.418}
&
\textbf{0.179}
&
+0.194
&
+0.083
\\

Honeycombing
&
0.015
&
0.007
&
\textbf{0.116}
&
\textbf{0.045}
&
+0.101
&
+0.038
\\

Seepage
&
0.186
&
0.099
&
\textbf{0.355}
&
\textbf{0.205}
&
+0.169
&
+0.106
\\

Rebar corrosion
&
0.071
&
0.026
&
\textbf{0.157}
&
\textbf{0.066}
&
+0.086
&
+0.040
\\

Rust
&
0.086
&
0.053
&
\textbf{0.171}
&
\textbf{0.100}
&
+0.085
&
+0.047
\\

Efflorescence
&
0.270
&
0.155
&
\textbf{0.436}
&
\textbf{0.250}
&
+0.166
&
+0.095
\\

\arrayrulecolor{black!22}
\midrule
\arrayrulecolor{black}

\rowcolor{summarygray}
\textbf{All}
&
\textbf{0.123}
&
\textbf{0.061}
&
\textbf{0.252}
&
\textbf{0.126}
&
\textbf{+0.129}
&
\textbf{+0.065}
\\

\bottomrule
\end{tabular*}

\vspace{1.2pt}

\begin{tablenotes}[flushleft]
\fontsize{5.85}{6.85}\selectfont
\item[]
Class rows report box AP, whereas the \textit{All} row reports box mAP averaged over the eight damage categories. Absolute gains are calculated as the enhanced score minus the corresponding low-light score. Bounding boxes were manually annotated on the original low-light images and reused for the enhanced counterparts because enhancement preserves spatial geometry. Both conditions use the same fixed YOLOv11m checkpoint and inference settings without enhancement-specific fine-tuning.
\end{tablenotes}

\end{threeparttable}
\endgroup

\vspace{-4pt}
\end{table}

As shown in Table~\ref{tab:detection_map}, DaL-MoE improves downstream box detection performance across all eight damage categories on the manually annotated real-world nighttime images. Compared with direct inference on the original low-light images, the overall box $\mathrm{mAP}_{50}$ increases from 0.123 to 0.252, representing an absolute gain of 0.129, while box $\mathrm{mAP}_{50{:}95}$ increases from 0.061 to 0.126, representing an absolute gain of 0.065. At the class level, the largest absolute $\mathrm{AP}_{50}$ gains are observed for hole, seepage, efflorescence, and spalling. Their $\mathrm{AP}_{50}$ values increase from 0.224, 0.186, 0.270, and 0.079 to 0.418, 0.355, 0.436, and 0.240, corresponding to absolute improvements of 0.194, 0.169, 0.166, and 0.161, respectively. Improvements are also obtained for challenging categories such as crack and honeycombing, whose $\mathrm{AP}_{50}$ values increase from 0.050 to 0.125 and from 0.015 to 0.116, respectively. A consistent improvement is observed in $\mathrm{AP}_{50{:}95}$ for every damage category, indicating that the recovered visual information benefits both damage recognition and localization under stricter IoU thresholds. Because the same fixed YOLOv11m checkpoint and inference settings are used for the low-light and enhanced images without enhancement-specific fine-tuning, these results provide quantitative real-scene evidence that DaL-MoE improves the usability of nighttime UAV bridge images for downstream damage detection.

Overall, the real-world evaluation supports the intended role of DaL-MoE as a detector-agnostic restoration front end for low-light UAV bridge inspection. The brightness and edge-density analyses demonstrate improved image-level visibility, while the qualitative examples and class-wise box detection results show that the recovered visual information can be utilized by a fixed downstream detector. Together with the full-reference restoration and multi-framework recognition results obtained on the paired synthetic benchmark, these findings provide complementary evidence of synthetic-to-real transfer. Broader validation on larger multi-site real-world datasets containing both bounding-box and instance-mask annotations will further establish robustness across bridge sites, camera systems, UAV platforms, and illumination conditions.

\section{Conclusions}
\label{sec:conclusions}

This paper addressed the degradation of UAV-based bridge damage recognition under low illumination by considering two related questions: how to establish controllable training and evaluation conditions without real paired normal-light--low-light images, and how to recover damage-related visual information that remains usable by different object detection and instance segmentation models. To establish controlled conditions, an ISP-aware synthesis pipeline was used to construct 2,895 spatially aligned normal-light--low-light image pairs while preserving the original bounding-box and instance-mask annotations for eight bridge damage categories. On this basis, the detector-agnostic DaL-MoE was developed to balance noise suppression, illumination and color adjustment, and structural-detail recovery through degradation-aware guidance and adaptive expert collaboration. Its standardized RGB output can be directly processed by existing recognition models without modifying their architectures or jointly training them with the restoration network.

On the held-out paired synthetic test set of 482 images, DaL-MoE achieved 23.1174~dB PSNR and 0.8482 SSIM, outperforming the compared low-light image restoration methods. These restoration improvements were also reflected in higher downstream recognition accuracy under the fixed-detector evaluation protocol. For YOLOv11m, box $\mathrm{mAP}_{50}$ increased from 0.3097 to 0.4923, while box $\mathrm{mAP}_{50{:}95}$ increased from 0.2105 to 0.3578. Mask $\mathrm{mAP}_{50}$ increased from 0.2281 to 0.3529, and mask $\mathrm{mAP}_{50{:}95}$ increased from 0.1051 to 0.1655. Similar improvements were observed for YOLOv8m, YOLOv8n, YOLOv11n, and SCNet, indicating that the restored images provide cross-architecture utility under the evaluated conditions. The ablation results further confirmed that degradation-aware guidance, the complementary expert branches, and adaptive gating each made a substantive contribution to the final restoration performance.

On 200 real-world nighttime UAV images without paired normal-light references, DaL-MoE increased the dataset-average grayscale intensity from 40.2 to 135.0 and the average Canny edge density from 1.55\% to 4.07\%. All 200 original low-light images were manually annotated with bounding boxes for the same eight damage categories used in the paired synthetic benchmark. Under the controlled protocol using the same fixed YOLOv11m checkpoint and inference settings without enhancement-specific fine-tuning, the overall box $\mathrm{mAP}_{50}$ increased from 0.123 to 0.252, corresponding to an absolute gain of 0.129, while box $\mathrm{mAP}_{50{:}95}$ increased from 0.061 to 0.126, corresponding to an absolute gain of 0.065. Class-wise $\mathrm{AP}_{50}$ and $\mathrm{AP}_{50{:}95}$ improved for all eight damage categories, and the qualitative examples likewise showed clearer damage regions and more complete predictions in representative scenes. Although changes in brightness and edge density are auxiliary visibility descriptors and do not directly measure structural fidelity, the controlled box detection results provide quantitative task-level evidence that the restored visual information can be utilized by a fixed downstream detector. Together, these image-level and task-level findings provide complementary qualitative and quantitative evidence of transfer from the ISP-aware synthetic degradation to real low-light bridge inspection scenes.

The conclusions of this paper should nevertheless be interpreted within the evaluated conditions. The three-channel RAW-like synthesis pipeline approximates exposure attenuation, spatially varying illumination, sensor noise, quantization, and ISP rendering, but it cannot fully reproduce the complexity of field imaging conditions, including more irregular illumination distributions, motion or defocus blur, and camera-dependent ISP and sensor-noise characteristics. The paired synthetic benchmark was derived from one collected bridge dataset acquired using a single imaging system, and only a limited set of downstream recognition architectures was evaluated. The current real-world quantitative evaluation is further limited to 200 images, bounding-box annotations, and one fixed YOLOv11m detector; it therefore does not yet establish real-scene instance segmentation performance or robustness across multiple detectors and independent inspection sites. In addition, a performance gap remains between the enhanced and normal-light inputs on the paired benchmark. DaL-MoE is primarily trained using an image-reconstruction objective and does not explicitly incorporate detection or segmentation supervision, leaving room for further improvement in downstream task performance. The results therefore indicate that DaL-MoE can serve as an effective low-light restoration front end under the tested conditions and exhibits promising cross-architecture and synthetic-to-real transfer potential, while broader independent validation remains necessary.

Future work will focus on strengthening the real-data foundation and extending the engineering applicability of the framework. On the data side, the current 200-image box-annotated real-world set will be expanded into a larger multi-site low-light bridge benchmark covering different bridge types, structural components, camera systems, UAV platforms, and illumination conditions. Complete bounding-box annotations will be retained and extended, while high-quality instance-mask annotations will be added to support quantitative real-scene evaluation of both object detection and instance segmentation across multiple recognition architectures. Real low-light observations can also be combined with camera calibration and RAW sensor information to improve the fidelity of degradation modeling and synthetic data generation. On the engineering side, future studies will investigate high-resolution bridge image processing, metric damage quantification, and spatial registration of damage predictions to bridge components. Integrating these results with BIM or digital-twin platforms, historical inspection records, and human verification workflows could further advance image-space damage recognition toward traceable component-level condition representation and maintenance decision support.


\section*{CRediT authorship contribution statement}
\noindent \textbf{Hu Wang:} Original Draft, Visualization, Validation, Coding, Funding Acquisition. 
\textbf{Hongxu Pu:} Validation, Coding, Data Curation.
\textbf{Zhiqi Hu:} Data Curation, Coding, Visualization. 
\textbf{Fangzhou Lin:} Data Curation, Figure Preparation. 
\textbf{Wang Wang:} Conceptualization, Original Draft, Experimental Design, Data Curation, Coding, Supervision.


\section*{Declaration of generative AI and AI-assisted technologies in the 
writing process}
\noindent During the preparation of this work the authors used GPT5.6 Sol in order to improve language. After using this tool, the authors reviewed and edited the content as needed and take full responsibility for the content of the publication.

\section*{Declaration of competing interest}
\noindent We declare that we do not have any commercial or associative interest that represents a conflict of interest in connection with the work submitted.

\section*{Data availability}
\noindent Data will be made available on request.

\bibliographystyle{my-elsarticle}
\bibliography{rmfile}

\end{document}